\documentclass[letterpaper]{article} % DO NOT CHANGE THIS
\usepackage[preprint]{aaai2027}  % DO NOT CHANGE THIS
\usepackage[hyphens]{url}  % DO NOT CHANGE THIS
\usepackage{graphicx} % DO NOT CHANGE THIS
\usepackage{natbib}  % DO NOT CHANGE THIS AND DO NOT ADD ANY OPTIONS TO IT
\usepackage{caption} % DO NOT CHANGE THIS AND DO NOT ADD ANY OPTIONS TO IT
\usepackage{algorithm}
\usepackage{algorithmic}
\usepackage{amsfonts}       % blackboard math symbols
\usepackage{amsmath}
\usepackage{makecell}
\usepackage{newfloat}
\usepackage{listings}
\DeclareCaptionStyle{ruled}{labelfont=normalfont,labelsep=colon,strut=off} % DO NOT CHANGE THIS
\floatstyle{ruled}
\newfloat{listing}{tb}{lst}{}
\floatname{listing}{Listing}

\usepackage{booktabs}
\usepackage{multirow}
\title{ProDVI: Programmatic Dynamics Priors for Value Network Initialization}
\author {
    Xinwei Liu\textsuperscript{\rm 1},
    Junyuan Liang\textsuperscript{\rm 1}\corresponding,
    Jianting Zhang\textsuperscript{\rm 2},
    Wuhui Chen\textsuperscript{\rm 1}
}
\affiliations {
    \textsuperscript{\rm 1}Sun Yat-sen University\\
    \textsuperscript{\rm 2}Purdue University\\
    liuxw73@mail2.sysu.edu.cn,
    liangjy53@mail2.sysu.edu.cn, 
    zhan4674@purdue.edu, chenwuh@mail.sysu.edu.cn
}

\begin{document}

\maketitle

\begin{abstract}
Deep Reinforcement Learning (RL) is notoriously sample inefficient.
One contributing factor is that RL agents are typically initialized from scratch, forcing them to acquire task-relevant knowledge through online interaction.
Existing approaches obtain informative initializations through pre-collected datasets, high-fidelity simulators, or meta-learning over related tasks, but these prerequisites may be difficult to access or even unavailable. 
In this paper, we propose Programmatic Dynamics Priors for Value Network Initialization (ProDVI), a framework that leverages the commonsense and domain knowledge encoded in large language models to initialize RL agents without relying on these resources.
Specifically, ProDVI prompts a code-generating language model to produce executable Python functions that encode coarse hypotheses about environment dynamics.
These functions are then used to generate synthetic transitions. Based on these transitions, we construct an auxiliary dynamics prediction objective to pretrain the state-action encoder of the value network in an actor-critic framework.
The learned representation provides dynamics-aware inductive biases before online RL begins.
Importantly, the generated programs are used only for representation pretraining and are not required to faithfully simulate the target environment. While the generated programs may be inaccurate, their induced initialization can be corrected through online learning from real transitions and rewards.
Experiments on OpenAI Gym and DeepMind Control Suite tasks show that ProDVI can effectively improve the sample efficiency of model-free RL algorithms.
\end{abstract}
\section{Introduction}
Deep Reinforcement Learning (RL) has achieved strong empirical performance across a wide range of decision-making problems \cite{DBLP:journals/nature/MnihKSRVBGRFOPB15, DBLP:journals/nature/SilverHMGSDSAPL16, hafner2025dreamerv3}, yet sample efficiency remains a central challenge \cite{DBLP:journals/corr/abs-1904-12901, DBLP:conf/icml/WangLYY024}. 
This inefficiency can be partly attributed to the fact that neural networks in RL are typically randomly initialized before training, without task-relevant inductive biases.
As a result, agents often have to learn task-relevant knowledge through interaction with the environment.
A long-standing goal in RL is therefore to obtain informative initializations that allow agents to adapt more quickly to target environments.
Existing approaches pursue this goal in different ways.
Offline-to-online RL methods first pretrain agents using previously collected trajectories and then fine-tune them through online interaction, thereby providing informative initializations for subsequent online RL \cite{DBLP:conf/nips/NakamotoZSM0FKL23, DBLP:conf/corl/RafailovHKMPF23, DBLP:conf/aaai/FengFSZL24}.
Sim-to-real methods initialize agents through training in simulation before transferring them to the target environment \cite{DBLP:conf/icra/PengAZA18, DBLP:conf/rss/KumarFPM21, DBLP:journals/scirobotics/RadosavovicXZDMS24}. Meta-RL methods learn task-adaptive initializations from a distribution of related tasks, enabling agents to adapt quickly to new environments with limited interaction \cite{DBLP:conf/icml/FinnAL17, DBLP:conf/iclr/RothfussLCAA19}. These approaches demonstrate that good initializations can effectively reduce the amount of online experience required for learning. However, they require pre-collected datasets, high-fidelity simulators, or a curated distribution of related tasks, which impose non-trivial barriers to acquiring informative initializations. This raises a question: 

\textit{Can we obtain informative initializations for RL agents without relying on these non-trivial prerequisites?}

Motivated by the broad commonsense and domain knowledge encoded in large language models (LLMs) \cite{DBLP:conf/iclr/HendrycksBBZMSS21, DBLP:journals/corr/abs-2303-08774, DBLP:journals/corr/abs-2601-03267}, in this work, we explore whether such knowledge can be used to provide informative parameter initializations for RL agents. This idea raises three design questions: 1) what kind of prior knowledge should be elicited from LLMs, 2) where this prior should be injected into an RL agent, and 3) how it should be distilled into the agent. 
To answer the first question, we prompt a code-generating LLM to generate prior knowledge about the dynamics of the environment in the form of executable Python functions. Given randomly sampled state-action pairs, the generated programs produce approximate next-state predictions. 
For the second question, we inject the generated dynamics-aware priors into the state-action encoder of the value network within an actor-critic RL framework \cite{sutton1998reinforcement}. 
For the third question, we first run the generated functions to produce a large number of synthetic transitions, and then construct an auxiliary task of dynamics prediction for the state-action encoder using the generated samples. In this way, prior knowledge about environment dynamics is distilled into an agent before online RL begins. We call the proposed method Programmatic Dynamics Priors for Value Network Initialization (ProDVI). 

ProDVI has four appealing properties. 
First, it does not require pre-collected trajectories, high-fidelity simulators, or meta-training tasks. The only external resources required by ProDVI are access to a code-generating LLM and metadata describing the target environment.
Second, because the generated dynamics functions are used only for dynamics-aware representation pretraining, they do not need to serve as faithful simulators of the target environment. 
As shown in our experiments, even imperfect programs can provide informative structural biases. 
During online RL, the value network is updated with real transitions and rewards, allowing inaccurate priors to be corrected by environment feedback.
Third, ProDVI decouples LLM usage from the online RL loop. Once the dynamics programs have been generated, the LLM is no longer queried, avoiding repeated LLM calls. 
Fourth, this form of prior enables reuse across tasks with shared dynamics. Because ProDVI focuses on dynamics priors rather than any specific task, the generated priors are specific to the underlying dynamics while remaining agnostic to individual tasks. Consequently, the same priors can apply to tasks governed by the same dynamics rules.

We evaluate ProDVI on tasks from OpenAI Gym and the DeepMind Control Suite. Experimental results demonstrate that ProDVI can effectively improve the sample efficiency of model-free RL algorithms. 

\section{Related Work}
\subsection{Informative Initializations for RL}
Previous work has investigated initializing RL agents in different ways. 
Offline-to-online RL methods obtain informative initializations by pretraining RL agents on previously collected trajectories and then fine-tuning them with online interaction \cite{DBLP:conf/nips/NakamotoZSM0FKL23, DBLP:conf/corl/RafailovHKMPF23, DBLP:conf/aaai/FengFSZL24}.
Sim-to-real methods initialize policies by training them in simulated environments before transferring them to the target domain \cite{DBLP:conf/icra/PengAZA18, DBLP:conf/rss/KumarFPM21, DBLP:journals/scirobotics/RadosavovicXZDMS24}.
Gradient-based meta-RL methods learn initial model parameters from a distribution of related tasks, enabling agents to rapidly adapt to new tasks with limited interaction \cite{DBLP:conf/icml/FinnAL17, DBLP:conf/iclr/RothfussLCAA19}.
These approaches demonstrate that informative initializations can reduce the amount of online experience required for learning.
However, they typically rely on pre-collected datasets, high-fidelity simulators, or curated task distributions. In contrast, ProDVI derives dynamics-aware priors from a code-generating LLM and uses them to pretrain the state-action encoder of the value network before online RL begins, which relaxes the prerequisites for obtaining informative RL initializations.

\subsection{LLM-Generated World Models and Simulations}
Recent work explores using LLMs to generate executable world models for decision-making. 
Code World Models \cite{DBLP:conf/nips/DaineseMAM24} use LLMs to generate Python programs that model environment dynamics.
Their method repeatedly queries the LLM to generate, improve, or fix candidate programs using feedback from unit tests and environment trajectories. The resulting code models are used for model-based planning. 
WorldCoder \cite{DBLP:conf/nips/0008KE24} iteratively builds a Python program as an executable world model through environment interaction and uses it for planning.
These methods are closely related to ProDVI because they also use LLM-generated code to capture environment dynamics. However, ProDVI does not aim to construct a faithful world model for planning. 
Instead, it uses the generated dynamics programs before online RL begins, producing synthetic transitions to pretrain the state-action encoder of the value network.
During online training, the generated programs are not used for planning or policy optimization, and the LLM is not queried again to refine them.

Another related direction uses generative models to expand the task and environment diversity for robot learning.
GenSim \cite{DBLP:conf/iclr/WangLYSBQWX024} uses LLMs to generate robotic simulation tasks and expert demonstrations for multitask policy learning. 
RoboGen \cite{DBLP:conf/icml/WangXCWWFEHG24} builds a generative pipeline that proposes robotic skills, constructs simulation environments, generates training supervision, and learns policies in the generated environments. 
Gen2Sim \cite{DBLP:conf/icra/KataraXF24} focuses on generating simulation assets, task descriptions, temporal task decompositions, and reward functions using language and vision generative models. These methods aim to expand simulation-based robot learning by generating richer tasks, environments, assets, or supervision. 
In contrast, ProDVI investigates how prior knowledge elicited from a code-generating LLM can be used to accelerate learning on a given task.

\subsection{LLM-Enhanced RL}
Recent work has explored using large language models (LLMs) to improve different components of RL. 
For reward design, Eureka \cite{DBLP:conf/iclr/MaLWHBJZFA24} prompts a code-generating LLM to produce executable reward functions and iteratively refine them using task feedback.
LORO \cite{DBLP:journals/corr/abs-2505-10861} warm-starts RL with LLM-generated off-policy data. 
LESR \cite{DBLP:conf/icml/Wang0JSLYJ24} uses LLM-generated code for task-relevant state engineering. 
LaRe \cite{DBLP:conf/aaai/0002JWMWLJ25} uses LLM-generated symbolic latent rewards to redistribute episodic returns and improve credit assignment in delayed-reward settings. 
LLM-Explorer \cite{DBLP:conf/nips/HaoSLYL25} periodically queries an LLM during training to analyze the agent's learning trajectory and adaptively guide subsequent policy exploration. 
ProDVI is orthogonal to these methods in terms of where the LLM-derived prior knowledge is applied. ProDVI targets network parameters rather than rewards, initial experience, input states, or exploration strategies.

\begin{figure*}[t]
\centering
\includegraphics[width=0.98\textwidth]{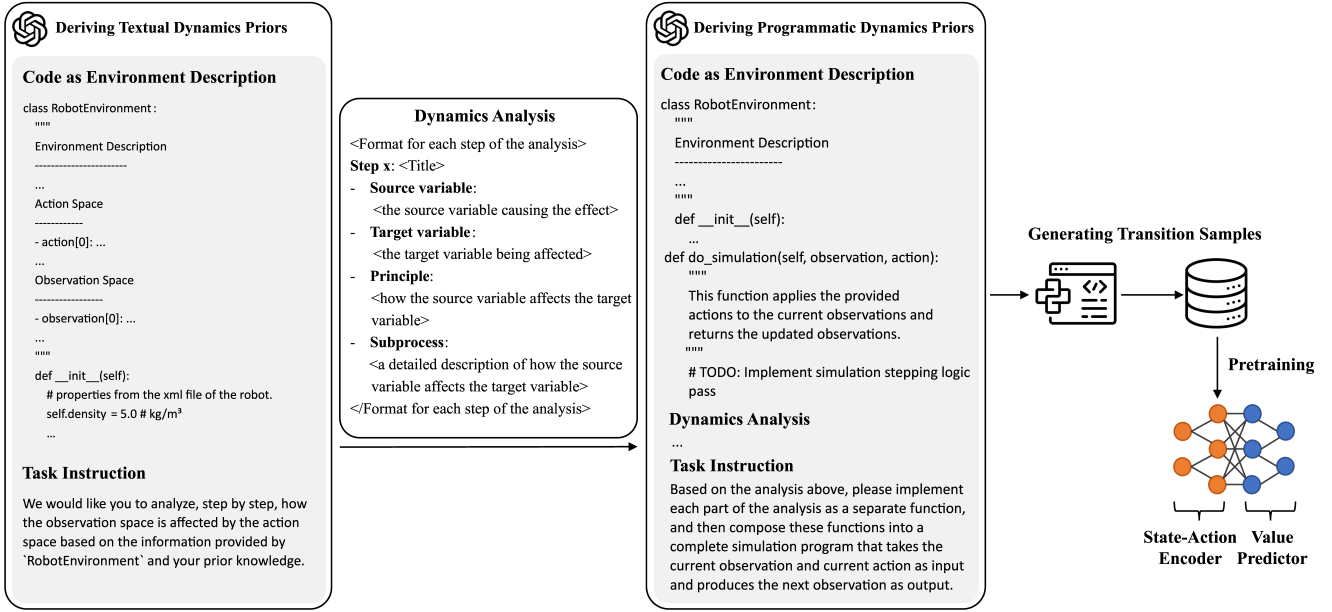} 
\caption{Overview of ProDVI. (1) Environment information is formatted as a Python class-style prompt for the LLM. The docstrings specify the action and observation spaces, while the internal variables store environment configuration parameters, if available, such as the robot's structural properties and the dimensions and densities of its components. (2) Given this Python-class description, an LLM first generates a textual dynamics analysis of how actions affect observations. (3) The LLM then implements the generated analysis as executable dynamics-prior code. (4) The generated code is used to produce synthetic transition samples, which are used to pretrain the state-action encoder of the value network.}
\label{framework}
\end{figure*}
\section{Preliminaries}
Reinforcement learning (RL) addresses the problem of sequential decision making, which is usually formulated as a Markov Decision Process (MDP). 
An MDP can be represented by a tuple $(\mathcal{S}, \mathcal{A}, P, r, \gamma)$, 
where $\mathcal{S}$ and $\mathcal{A}$ denote the state and action spaces, respectively; $P(s_{t+1}|s_t,a_t)$ denotes the transition probability of the next state \begin{math} s_{t+1} \end{math} given the current state \begin{math} s_t \end{math} and action $a_t$; $r:\mathcal{S} \times \mathcal{A} \rightarrow \mathbb{R}$ is the reward function; $\gamma \in [0, 1)$ is the discount factor. The objective of RL is to learn a policy $\pi:\mathcal{S} \rightarrow \mathcal{A}$ that maximizes the discounted cumulative reward $\sum_{t=0}^{\infty} \gamma^{t} r_{t}$. Actor-critic methods \cite{sutton1998reinforcement} typically learn an action-value function $Q^{\pi}(s,a)= \mathbb{E}_{\pi}[\sum_{t=0}^{\infty} \gamma^{t} r_{t}|s_0=s,a_0=a]$ which models the expected return, starting from an initial state \begin{math}s\end{math} and action \begin{math}a\end{math}. 
In deep actor-critic methods \cite{DBLP:conf/icml/FujimotoHM18, DBLP:conf/icml/HaarnojaZAL18}, the action-value function is usually approximated by a neural network, often referred to as a value network or critic. Given a state-action pair $(s,a)$, the value network first maps the input into a latent representation and then predicts the corresponding action value. We refer to the representation-learning component of the value network as the state-action encoder. 

In this paper, we focus on state-based settings, where the observation received by the agent at time step $t$, denoted by $o_t$, can typically be treated as the Markovian state $s_t$. Therefore, we slightly abuse the terminology of state and observation, and use $o_t$ and $s_t$ interchangeably when the context is clear.

\section{Method}
This section introduces ProDVI by addressing the three design questions raised in the introduction. Section~\ref{sec:derive-priors} describes what form of prior knowledge is derived from an LLM, while Section~\ref{sec:prior-distillation} explains where this prior is injected into the RL agent and how it is incorporated before online RL begins. Figure~\ref{framework} demonstrates the overview of ProDVI.

\subsection{Deriving Dynamics Priors from LLMs}
\label{sec:derive-priors}
Informative priors play a crucial role in improving the performance and sample efficiency of RL agents. The key challenge, however, is to identify what form of prior knowledge can provide informative guidance for learning. 
Prior work suggests that dynamics-aware representations can significantly improve the sample efficiency and performance of model-free methods \cite{DBLP:conf/icml/OtaOJMN20, anonymous2026preliminary}. 
For example, OFENet \cite{DBLP:conf/icml/OtaOJMN20} trains state-action representations by predicting the next observation, thereby encouraging the state-action encoder to capture dynamics knowledge about the environment.
Denoting $f_{\theta}$ as a state-action encoder parameterized by $\theta$, and given a transition sample $(o_t, a_t, o_{t+1})$, $f_{\theta}$ is optimized with the following auxiliary loss:
\begin{equation}
\label{eq:1}
\begin{aligned}
z_t &= f_{\theta}\left(o_t, a_t\right) \\
\mathcal{L}_\text{Aux}&= \left\|{d_{\phi}}(z_t) - o_{t+1}\right\|_2^2,
\end{aligned}
\end{equation}
where $z_t$ represents a state-action representation, and ${d_{\phi}}$ denotes a decoder with parameters $\phi$. The state-action representation is taken as the input to a value predictor for value learning in RL. Recent work further suggests that this auxiliary dynamics prediction task should be performed in a normalized observation space \cite{anonymous2026preliminary}. 
Since different observation dimensions may have substantially different value ranges, directly predicting raw observations can lead to imbalanced gradients across dimensions, encouraging the learned representations to neglect dimensions with relatively small ranges \cite{anonymous2026preliminary}. 
The Equation~\ref{eq:1} is thus modified as follows:
\begin{equation}
\label{eq:1.1}
\begin{aligned}
z_t &= f_{\theta}\left(\tilde{o}_t, a_t\right) \\
\mathcal{L}_\text{Aux}&= \left\|{d_{\phi}}(z_t) - \tilde{o}_{t+1}\right\|_2^2,
\end{aligned}
\end{equation}
where $\tilde{o}_t$ and $\tilde{o}_{t+1}$ denote the normalized current and next observations, respectively. In \cite{anonymous2026preliminary}, streaming observations are normalized by a method tailored to the characteristics of online RL.

Inspired by these dynamics-based representation learning methods, we posit that prior knowledge about environment dynamics can provide informative initializations for RL agents. 
Given the extensive commonsense and domain knowledge encoded in LLMs, we use an LLM to derive dynamics-aware priors for RL. 
Similar to Eureka \cite{DBLP:conf/iclr/MaLWHBJZFA24}, we describe the environment in a code-based format. However, ProDVI does not require access to the executable environment implementation. 
Instead, it uses lightweight environment information to construct a Python-class description.
The docstrings of this class describe the action and observation spaces, including the semantic meaning of each dimension. The class attributes store environment configurations, such as gravity and body masses, when available.

As illustrated in Figure~\ref{framework}, given the environment description, an LLM is prompted to generate textual prior knowledge about the target environment dynamics. This textual dynamics prior provides a structured analysis of how the current observation is affected by the action and how the current observation evolves into the next observation step by step.
Based on the environment description and the LLM-generated textual dynamics analysis, ProDVI further prompts the LLM to implement the analysis as executable Python functions. This yields an approximate dynamics-prior program, denoted by $g_{\mathrm{LLM}}$, which maps an observation-action pair to the next observation.
$g_{\mathrm{LLM}}$ is not intended to be a faithful simulator for planning. Instead, it serves as a source of approximate dynamics priors for representation pretraining.

\subsection{Dynamics Prior Distillation}
\label{sec:prior-distillation}
We inject programmatic dynamics priors into the state-action encoder of the value network. This design connects ProDVI to previous dynamics-based representation learning methods. ProDVI can be viewed as an LLM-enhanced representation learning method that does not require real environment transitions before online RL begins. Following \cite{anonymous2026preliminary}, ProDVI performs representation learning and online RL in a normalized observation space.

To distill the LLM-generated dynamics priors into the state-action encoder, we first generate synthetic observation-action pairs.
Specifically, observations are sampled from a multivariate Gaussian distribution and clipped to finite ranges to avoid extreme values, while actions are uniformly sampled from their lower and upper bounds:
\begin{equation}
\label{eq:2}
\begin{gathered}
o \sim \mathcal{N}(0, I), \quad \tilde{o} = \operatorname{clip}(o, -O, O), \\ {a} \sim \mathcal{U}(a_{\min}, a_{\max}).
\end{gathered}
\end{equation}
We then feed the sampled observation-action pairs into the generated dynamics-prior program ${g}_{\mathrm{LLM}}$ to obtain the pseudo next observations:
\begin{equation}
\label{eq:4}
\hat{o}_{t+1} = g_{\mathrm{LLM}}(\tilde{o}_t, {a}_t).
\end{equation}
Using Equation~\ref{eq:4}, we generate a set of input-output pairs.
Since the inputs $(\tilde{o}_t,a_t)$ are randomly sampled, they may not lie on the real observation-action manifold.
However, our goal is not to generate samples from the true observation-action distribution.
We aim to construct a diverse set of input-output pairs generated by $g_{\mathrm{LLM}}$, so that they can be used to train a neural network to approximate $g_{\mathrm{LLM}}$.

Before pretraining, we normalize the generated next observations $\hat{o}_{t+1}$ using the per-dimension mean and standard deviation computed over the synthetic dataset, and then clip the normalized values to a finite range, yielding $\tilde{o}_{t+1}$. 
Thus, the pretraining target corresponds to an affine transformation of the outputs of $g_{\mathrm{LLM}}$, followed by clipping.
We finally pretrain the state-action encoder using $(\tilde{o}_t,a_t,\tilde{o}_{t+1})$ with the auxiliary loss in Equation~\ref{eq:1.1}, obtaining the state-action encoder parameters $\bar{\theta}$. 
Although the dynamics prior distilled into the encoder may not accurately reflect the true dynamics in the normalized observation space, it can be subsequently corrected through online updates using real transitions.

We empirically find that simply using the pretrained parameters $\bar{\theta}$ as the state-action encoder initialization does not always yield the best RL performance. One possible explanation is plasticity loss in deep neural networks, which refers to the gradual deterioration of a network’s ability to learn from new data after extensive training \cite{DBLP:conf/collas/AbbasZM0M23, DBLP:journals/nature/DohareHLRMS24}.
To mitigate this issue, we apply a soft reset to $\bar{\theta}$, following the shrink-and-perturb strategy adopted by \cite{DBLP:conf/iclr/DOroSNBBC23}:
\begin{equation}
\theta_{\mathrm{init}} = \alpha \bar{\theta} + (1-\alpha)\theta_0,
\end{equation}
where $\theta_0$ denotes a new set of parameters sampled from the network’s original initialization distribution, and $\alpha \in (0,1)$ controls the extent to which the previously learned parameters are retained.

\section{Experiments}
In our experiments, we aim to answer the following questions:
\begin{enumerate}
    \item Can ProDVI improve the sample efficiency and performance of strong model-free RL algorithms?
    \item Are the LLM-generated dynamics priors transferable to other RL algorithms?
    \item Why do the LLM-generated dynamics priors improve RL?
    \item Does ProDVI exhibit robustness across the dynamics priors generated by different runs of LLM?
\end{enumerate}

\subsection{Experimental Setup}
\paragraph{Environments.} We evaluate ProDVI on two widely used state-based continuous-control benchmarks, OpenAI Gym \cite{DBLP:conf/nips/TowersKBCDGKKKP25} and the DeepMind Control Suite (DMControl) \cite{DBLP:journals/corr/abs-1801-00690}.
We consider 5 common locomotion tasks from Gym, and 7 challenging tasks from the dog and humanoid domains in DMControl.
For Gym tasks, agents are trained for 1M time steps. For DMControl tasks, agents are trained for 500k time steps, equivalent to 1M frames in the original environment due to an action repeat of 2.

\begin{table*}[t]
    % \small
    \centering
    \begin{tabular}{cccccc}
        \toprule
        \textbf{\makecell{Environment\\Steps}}  & \textbf{Metrics} & \textbf{\makecell{AnonMethod\\w/o Aux}} & \textbf{\makecell{AnonMethod \\ w/o Aux + ProDVI}} & \textbf{AnonMethod} & \textbf{\makecell{AnonMethod\\+ ProDVI}} \\
        \midrule
        \multicolumn{6}{l}{\textit{Gym}} \\
        \midrule        
        \multirow{2}{*}{0.25M}
        &Mean &
        0.87 [0.77, 0.98] & 
        \textbf{{1.20}} [1.12, 1.28] &
        1.05 [0.93, 1.17] & 
        \textbf{{1.33}} [1.24, 1.42] \\

        &IQM & 
        0.79 [0.70, 0.89] & 
        \textbf{1.11} [1.01, 1.17] & 
        0.88 [0.76, 1.01] & 
        \textbf{1.15} [1.03, 1.25] \\
        
        \midrule

        \multirow{2}{*}{0.50M}
        &Mean & 
        1.36 [1.30, 1.43] & 
        \textbf{{1.58}} [1.52, 1.65] & 
        1.45 [1.32, 1.55] & 
        \textbf{{1.64}} [1.57, 1.71] \\
        
        &IQM & 
        1.15 [1.08, 1.20] & 
        \textbf{{1.33}} [1.24, 1.40] & 
        1.16 [0.98, 1.30] & 
        \textbf{{1.40}} [1.34, 1.45] \\
        
        \midrule    

        \multirow{2}{*}{1.00M}
        &Mean & 
        1.55 [1.47, 1.62] &  
        \textbf{{1.76}} [1.69, 1.83] & 
        1.78 [1.74, 1.83] &  
        \textbf{{1.85}} [1.77, 1.92] \\

        &IQM & 
        1.30 [1.20, 1.41] & 
        \textbf{{1.50}} [1.43, 1.56] & 
        1.50 [1.44, 1.56] & 
        \textbf{{1.54}} [1.48, 1.63] \\
    
        \midrule
        \multicolumn{4}{l}{\textit{DMControl}} \\
        \midrule     
        \multirow{2}{*}{0.25M}
        &Mean & 359 [342, 375] & \textbf{396} [361, 430] & 394 [373, 416] & \textbf{433} [417, 453] \\
        &IQM & 308 [286, 334] & \textbf{351} [312, 393] & 343 [320, 369] & \textbf{392} [374, 411] \\
        
        \midrule

        \multirow{2}{*}{0.50M}
        &Mean & 589 [564, 614] & \textbf{620} [586, 654] & 628 [602, 649] & \textbf{646} [628, 662] \\
        &IQM & 608 [567, 648] & \textbf{653} [593, 712] & 680 [634, 718] & \textbf{701} [668, 728] \\      
        \bottomrule
    \end{tabular}
    \caption{
    Aggregated scores on Gym and DMControl at different environment-step budgets. Gym scores are Deep-TD3-normalized, while DMControl scores are raw episode returns.  
    Bold numbers indicate the better performance under the same metric and environment-step budget. Brackets denote 95\% bootstrap confidence intervals. Full per-task results are provided in the appendix.
    }
    \label{main_result}
\end{table*}

\paragraph{Implementation Details.} We employ GPT-5.5 as the LLM used by ProDVI. The prompt templates and details of prompts are available in the appendix. For Gym, ProDVI generates task-specific dynamics priors. For DMControl, however, ProDVI generates dynamics priors at the domain level (i.e., dog and humanoid domains), since tasks within the same domain, such as dog-$\{\text{trot, run, stand, walk}\}$, share the same underlying dynamics rules.

ProDVI adopts AnonMethod \cite{anonymous2026preliminary}, a recent model-free RL method built on TD3 \cite{DBLP:conf/icml/FujimotoHM18}, as the backbone RL algorithm because it outperforms state-of-the-art model-free \cite{DBLP:conf/nips/FujimotoCSGPM23, DBLP:conf/iclr/FujimotoD0TR25} and model-based RL methods \cite{DBLP:conf/iclr/00010024, hafner2025dreamerv3} on Gym and DMControl. 
In state-based settings, AnonMethod learns state-action representations by coupling value learning with an auxiliary task that predicts the next normalized observation. Both the input and target observations are normalized by NormMethod, a normalization method designed for state-based online RL, to balance the auxiliary losses across observation dimensions. A detailed description of AnonMethod and NormMethod is provided in the appendix.
We instantiate the observation normalization required by ProDVI with NormMethod, so that online observations are kept on a scale comparable to the synthetic observations used for pretraining.
We use the default hyperparameters of AnonMethod and NormMethod.
When combined with AnonMethod, ProDVI distills the LLM-generated dynamics priors into AnonMethod's state-action encoder before online RL. 
When pretraining, synthetic observations are normalized using statistics computed over the full synthetic transition dataset. During online RL, NormMethod is used to keep real observations on a comparable normalized scale. 
% Additional details are provided in the appendix.

ProDVI uses $7$M synthetic transitions and sets $\alpha=0.5$ as the default configuration for all experiments. This configuration is selected through a simple two-stage sensitivity study on Gym. We first fix $\alpha=1.0$ and vary the number of synthetic transitions among ${1,3,5,7,10}$M, where $7$M achieves the best overall performance. We then fix the number of synthetic transitions to $7$M and vary $\alpha$ among $\{1.0,0.8,0.5,0.2\}$, finding that $\alpha=0.5$ provides strong performance. This procedure is not an exhaustive grid search over all hyperparameter combinations. Once selected, the same default configuration is used for all subsequent experiments, including those on DMControl. The full sensitivity results are provided in the appendix.

\paragraph{Evaluation Protocol.} All experiments are run for 5 seeds. For each seed, we evaluate the agent every 5k environment steps over 10 episodes and report the average episode return as the evaluation score. 

The maximum total return for each episode of DMControl tasks is 1000. For Gym tasks, following AnonMethod, we normalize the score of each task by the performance of a \textit{deep} variant of TD3 before aggregating results across tasks:
{\small
\begin{equation}
\label{eq:3}
\begin{aligned}
\operatorname{Deep-TD3-Norm}(x) = \frac{x-\text{random score}}{{\text{Deep-TD3 score} - \text{random score}}},
\end{aligned}
\end{equation}}
where $x$ denotes the evaluation return on a given task.
The Deep-TD3 baseline increases the number of linear layers and the hidden size of vanilla TD3, so that its number of trainable parameters is comparable to that of AnonMethod.
This provides a stronger and more size-matched reference point for normalization than vanilla TD3. See the appendix for the implementation details.
We report the mean and interquartile mean (IQM) across tasks as aggregate metrics.

\subsection{ProDVI Improves Sample Efficiency}
\begin{table*}[t]
    \centering
    \begin{tabular}{cccccc}
        \toprule
        \textbf{\makecell{Environment\\Steps}}  & \textbf{Metrics} & \textbf{\makecell{AnonMethod-SAC \\ w/o Aux}} & \textbf{\makecell{AnonMethod-SAC \\ w/o Aux + ProDVI}} & \textbf{\makecell{AnonMethod-SAC}} & \textbf{\makecell{AnonMethod-SAC\\+ ProDVI}}  \\
        \midrule     
        \multirow{2}{*}{0.25M}
        &Mean &
        0.54 [0.43, 0.66] & 
        \textbf{0.68} [0.61, 0.76] &
        0.62 [0.53, 0.71] & 
        \textbf{0.98} [0.83, 1.11] \\

        &IQM & 
        0.55 [0.41, 0.67] & 
        \textbf{0.68} [0.60, 0.78] &
        0.67 [0.56, 0.76] & 
        \textbf{0.90} [0.80, 1.09] \\
        
        \midrule

        \multirow{2}{*}{0.50M}
        &Mean & 
        0.98 [0.85, 1.10] & 
        \textbf{1.04} [0.88, 1.19] &
        0.94 [0.83, 1.04] & 
        \textbf{{1.17}} [1.06, 1.29] \\
        
        &IQM & 
        0.97 [0.85, 1.09] & 
        \textbf{1.01} [0.89, 1.14] &
        0.96 [0.83, 1.06] & 
        \textbf{{1.10}} [1.01, 1.22] \\
        
        \midrule    

        \multirow{2}{*}{1.00M}
        &Mean & 
        1.10 [0.95, 1.23] &  
        \textbf{1.20} [1.07, 1.31] &
        1.28 [1.21, 1.34] & 
        \textbf{{1.34}} [1.23, 1.46] \\

        &IQM & 
        1.07 [0.91, 1.22] & 
        \textbf{{1.20}} [1.13, 1.27] &
        1.26 [1.21, 1.33] & 
        \textbf{{1.27}} [1.18, 1.38] \\
      
        \bottomrule
    \end{tabular}
    \caption{
    Aggregated Deep-SAC-normalized scores on Gym at different environment-step budgets. Bold numbers indicate the better result within each with/without-ProDVI pair under the same metric and environment-step budget. Brackets denote 95\% bootstrap confidence intervals. Full per-task results are provided in the appendix.
    }
    \label{main_result3}
\end{table*}

Table~\ref{main_result} compares the aggregate performance of AnonMethod with and without ProDVI on Gym and DMControl.
The results show that ProDVI can further improve the overall performance of AnonMethod, a strong model-free observation-predictive method, on these benchmarks. 
Since AnonMethod learns dynamics-aware representations through an online observation-prediction objective, these gains suggest that ProDVI is not merely compensating for the absence of representation learning. Instead, it provides an informative initialization before real environment interaction, thereby reducing the amount of online experience required for the value network to acquire informative state-action features.
The improvement is especially pronounced in the low-data regime, for example at 0.25M environment steps, which indicates that ProDVI can effectively improve sample efficiency.

We also consider AnonMethod w/o Aux, a variant constructed by removing the observation prediction task from AnonMethod. 
This variant can be viewed as a TD3 variant enhanced with a set of lightweight techniques. A detailed comparison between AnonMethod w/o Aux and vanilla TD3 is provided in the appendix.
As shown in Table~\ref{main_result}, ProDVI consistently improves the aggregate performance of AnonMethod w/o Aux on Gym across all reported environment-step budgets.
This suggests that the benefit of ProDVI is not contingent on whether an auxiliary observation-prediction task is used.

\subsection{Transferring to Other Algorithms}
To examine whether ProDVI is tied to a specific RL backbone, we apply the same programmatic dynamics priors to a SAC-based variant of AnonMethod, denoted as AnonMethod-SAC.
This variant replaces the TD3 backbone used in AnonMethod with SAC \cite{DBLP:conf/icml/HaarnojaZAL18}.
We also consider AnonMethod-SAC w/o Aux, which can be viewed as an enhanced SAC baseline equipped with the same lightweight techniques, but without the auxiliary task of observation prediction.
The detailed differences between AnonMethod-SAC w/o Aux and vanilla SAC are provided in the appendix.
Table~\ref{main_result3} reports the aggregated Deep-SAC-normalized scores on Gym. Deep-SAC-normalized scores are computed in the same way as the Deep-TD3-normalized scores in Equation~\ref{eq:3}, except that the Deep-TD3 score is replaced by that of a \textit{deep} variant of SAC (see the appendix).
ProDVI improves both AnonMethod-SAC w/o Aux and AnonMethod-SAC across different environment-step budgets, with particularly large gains early in training. 
These suggest that the dynamics priors learned from LLM-generated programs are not specialized to TD3-based methods.

\begin{table}[t]
    % \small
    % \setlength{\tabcolsep}{1mm}
    \centering
    \begin{tabular}{llccc}
        \toprule
        \multirow{2}{*}{\textbf{Task}} & \multirow{2}{*}{\textbf{Loss Type}}  & \multicolumn{3}{c}{\textbf{Aggregated Metrics}}  \\
        \cmidrule(lr){3-5}
        & & \textbf{Mean} & \textbf{Std} & \textbf{IQM} \\
        \midrule        
        \multirow{2}{*}{Ant} & MSE &1.4e4 & 3.7e4 & 980 \\
         & MAPE (\%) &8.3e4 & 8.9e6 & 1645 \\
        \midrule 
        \multirow{2}{*}{HalfCheetah} & MSE & 313 & 255 & 264 \\
         & MAPE (\%) &3036 & 1.1e5 & 801 \\
        \midrule 
        \multirow{2}{*}{Hopper} & MSE & 0.31 & 0.76 & 0.18 \\
         & MAPE (\%) &1.0e4 & 4.5e5 & 59 \\
        \midrule 
        \multirow{2}{*}{Humanoid} & MSE & 2807 & 4233 & 1673 \\
         & MAPE (\%) &2.9e7 & 2.9e7 & 2.2e7 \\
        \midrule 
        \multirow{2}{*}{Walker2d} & MSE & 8 & 8 & 6 \\
         & MAPE (\%) &5.2e4 & 4.5e6 & 74 \\

        \bottomrule
    \end{tabular}
    \caption{
    Prediction errors of LLM-generated dynamics-prior programs on Gym tasks. The table reports aggregated MSE and MAPE between the predicted next observations and the true next observations. 
    }
    \label{loss_result}
\end{table}

\subsection{How LLM-Generated Dynamics Priors Improve RL}
ProDVI significantly improves the sample efficiency and performance of model-free RL methods. A natural explanation is that the LLM-generated dynamics-prior programs may accurately reflect the true environment dynamics, and this accurate dynamics knowledge directly benefits RL agents. To examine this hypothesis, we evaluate whether the LLM-generated dynamics-prior programs can accurately predict the next observation on the Gym benchmark. 
% The details are provided in the appendix.
Table~\ref{loss_result} reports the aggregated prediction errors of the LLM-generated dynamics-prior programs on the 5 Gym tasks. 
For each task, the evaluation data consist of 100k transition samples collected by interacting with the environment using randomly sampled actions. 
The results show that the dynamics encoded by these programs are far from the true environment dynamics. This finding rules out the accuracy-based explanation. 

We therefore hypothesize that these programs may contain information that is informative for learning representations for dynamics modeling. To verify this hypothesis, we consider a dynamics prediction task, where a predictor takes the current action and normalized observations as inputs and predicts the next normalized observation. 
We take the state-action encoder and decoder obtained after dynamics priors distillation as a dynamics predictor, reset the last layer of the decoder, and then train the predictor using transition data sampled from the replay buffer of Deep-TD3.
We then report the training losses after 0.25M and 0.50M gradient updates. As a baseline, we train another predictor with the same architecture from scratch using the same sampled data. Experimental details are provided in the appendix.
Table~\ref{probe_result} summarizes the results on Gym. Overall, the initialization obtained after dynamics priors distillation leads to lower training losses than training from scratch in most tasks and update budgets. This suggests that although the LLM-generated dynamics priors are not accurate dynamics models, they still contain informative information for dynamics modeling.

\begin{table}[t]
    \setlength{\tabcolsep}{1mm}
    \centering
    \begin{tabular}{lcc}
        \toprule
        {\textbf{Task}}
        & \textbf{From Scratch} & \textbf{Dynamics Priors}  \\
        \midrule  
        \multicolumn{3}{l}{\textit{0.25M Gradient Updates}}\\
        \midrule
        {Ant}  &9.9e-3 [9.8, 10]e-3 & \textbf{9.5e-3} [9.2, 9.8]e-3 \\      
        {HalfCheetah}  &7.4e-3 [6.9, 8.2]e-3 & \textbf{6.3e-3} [6.0, 6.8]e-3 \\
        {Hopper}  &3.3e-4 [2.9, 3.8]e-4 & \textbf{2.5e-4} [2.3, 2.7]e-4 \\
        {Humanoid}  &0.055 [0.054, 0.057] & \textbf{0.053} [0.051, 0.054] \\
        {Walker2d}  &\textbf{5.9e-3} [5.7, 6.1]e-3 & 6.0e-3 [5.7, 6.2]e-3 \\
        \midrule  
        \multicolumn{3}{l}{\textit{0.50M Gradient Updates}}\\
        \midrule  
        {Ant}  &8.1e-3 [7.9, 8.3]e-3 & \textbf{7.6e-3} [7.4, 7.7]e-3 \\      
        {HalfCheetah} & 4.5e-3 [4.2, 4.8]e-3 & \textbf{4.2e-3} [3.9, 4.7]e-3 \\
        {Hopper}  &1.8e-4 [1.7, 2.0]e-4 & \textbf{1.7e-4} [1.6, 1.8]e-4 \\
        {Humanoid}  &0.042 [0.041, 0.044] & 0.042 [0.040, 0.044] \\
        {Walker2d}  &4.4e-3 [4.3, 4.6]e-3 & \textbf{4.3e-3} [4.2, 4.4]e-3 \\       
        \bottomrule
    \end{tabular}
    \caption{
    Training losses of dynamics predictors on transition data collected by Deep-TD3 agents. 
    Bold numbers indicate lower losses under the same task and update budget. Brackets denote 95\% bootstrap confidence intervals over 5 seeds.
    }
    \label{probe_result}
\end{table}

\begin{table}[t]
    \small
    \setlength{\tabcolsep}{1mm}
    \centering
    \begin{tabular}{ccccc}
        \toprule
        \textbf{\makecell{Environment\\Steps}}  & \textbf{Metrics} & \textbf{\makecell{Dynamics\\Prior \# 2 }} & \textbf{\makecell{Dynamics\\Prior \# 3}}  \\
        \midrule  
        \multicolumn{4}{l}{\textit{Gym}} \\
        \midrule     

        \multirow{2}{*}{0.25M}
        &Mean &
        1.19 [1.07, 1.31] & 
        {1.24} [1.14, 1.34] \\

        &IQM & 
        1.04 [0.92, 1.13] & 
        {1.05} [0.91, 1.15] \\
        
        \midrule

        \multirow{2}{*}{0.50M}
        &Mean & 
        1.58 [1.50, 1.66] & 
        {{1.58}} [1.49, 1.67] \\
        
        &IQM & 
        1.32 [1.22, 1.40] & 
        {{1.27}} [1.16, 1.38] \\
        
        \midrule    

        \multirow{2}{*}{1.00M}
        &Mean & 
        1.82 [1.76, 1.88] & 
        {{1.81}} [1.77, 1.86] \\

        &IQM & 
        1.52 [1.47, 1.58] & 
        {{1.52}} [1.46, 1.58] \\
        \midrule  
        \multicolumn{4}{l}{\textit{DMControl}} \\
        \midrule  
        \multirow{2}{*}{0.25M}
        &Mean &
        438 [420, 457] & 
        407 [389, 424] \\

        &IQM & 
        410 [390, 434] & 
        369 [340, 397] \\
        
        \midrule

        \multirow{2}{*}{0.50M}
        &Mean & 
        651 [598, 690] & 
        652 [634, 670] \\
        
        &IQM & 
        723 [657, 771] & 
        712 [682, 741] \\
        
        \bottomrule
    \end{tabular}
    \caption{
    Aggregated scores of AnonMethod with ProDVI using two independently generated dynamics-prior programs.
    }
    \label{robustness}
\end{table}

\subsection{Robustness to Different LLM-Generated Dynamics Priors}
Since LLM-generated programs are stochastic, a natural question is whether ProDVI is sensitive to a particular generated prior.
Starting from the same Python-class description, we make two additional independent calls to the LLM and obtain different textual dynamics analyses and executable dynamics-prior programs. We refer to the resulting programs as \textbf{Dynamics Prior \#2} and \textbf{Dynamics Prior \#3}.
For each generated dynamics-prior program, we independently generate synthetic transitions and pretrain a separate initialization for the state-action encoder of the value network, while keeping all other pretraining and online RL settings unchanged.

Table~\ref{robustness} reports the aggregated performance of AnonMethod initialized with the parameters induced by these two dynamics-prior programs. 
Both dynamics priors improve the aggregate performance on Gym and DMControl.
These results suggest that ProDVI is robust to the stochasticity of LLM generation and remains effective when using dynamics priors generated from different runs.

\section{Conclusion and Limitations}
In this paper, we proposed Programmatic Dynamics Priors for Value Network Initialization (ProDVI), a framework that uses LLM-generated programmatic dynamics priors to provide informative initializations for RL agents.
Unlike offline-to-online RL, sim-to-real transfer, and meta-RL methods, ProDVI does not require pre-collected trajectories, high-fidelity simulators, or curated task distributions.
Instead, it uses environment descriptions to prompt a code-generating LLM to produce approximate dynamics programs, which generate synthetic transitions for pretraining the state-action encoder of the value network within an actor-critic framework.
Experiments on OpenAI Gym and DeepMind Control Suite show that ProDVI improves the sample efficiency and performance of strong model-free RL algorithms. Experiment results suggest that its gains come not from accurate dynamics simulation, but from informative dynamics information distilled into hidden units. Together, these demonstrate that programmatic dynamics priors provide a promising way to use LLM knowledge for sample-efficient RL.

ProDVI has several limitations.
First, ProDVI is currently designed for state-based settings. Extending ProDVI to more complex settings, such as high-dimensional visual observations, remains an important direction for future work.
Besides, although ProDVI does not require the LLM-generated programs to faithfully simulate the target environment, the quality of the generated dynamics priors is still constrained by the capabilities of the LLM. These programs are not guaranteed to capture informative task-relevant dynamics regularities. This limitation is expected to be mitigated as LLMs continue to improve in code generation, physical reasoning, and domain-specific understanding.

{
\small
\bibliography{aaai2027}}
\end{document}

\twocolumn[
\begin{center}
\section*{Supplementary Material}
\end{center}
]

\appendix

\section{Pseudocode}
Algorithm~\ref{algorithm:prodvi} summarizes the ProDVI pipeline.

\begin{algorithm}[tb]
\caption{Programmatic Dynamics Priors for Value Network Initialization}
\label{algorithm:prodvi}

Let $\mathcal{E}$ denote the code-style environment description, $\mathcal{M}$ denote a code-generating LLM.

Let $N_{\mathrm{syn}}$ denote the number of synthetic transitions generated for dynamics-prior pretraining.

Let $f_{\theta}$ denote the state-action encoder of the value network, $d_{\phi}$ denote the decoder for dynamics-prior pretraining, and $\alpha \in (0,1)$ denote the interpolation coefficient.

\begin{algorithmic}[1]
\STATE Prompt $\mathcal{M}$ with $\mathcal{E}$ to generate textual dynamics analysis and executable dynamics-prior program ${g}_{\mathrm{LLM}}$
\STATE Initialize an empty synthetic dataset $\mathcal{D}_{\mathrm{syn}}$
\FOR{$i=1$ to $N_{\mathrm{syn}}$}
    \STATE Sample $o_i \sim \mathcal{N}(0,I)$ and set $\tilde{o}_i=\operatorname{clip}(o_i,-O,O)$
    \STATE Sample an action uniformly from the valid action range $a_i \sim \mathcal{U}(a_{\min},a_{\max})$
    \STATE Compute $\hat{o}_{i+1} = {g}_{\mathrm{LLM}}(\tilde{o}_i,{a}_i)$ 
    \STATE Add $(\tilde{o}_i,{a}_i,\hat{o}_{i+1})$ to $\mathcal{D}_{\mathrm{syn}}$
\ENDFOR
\STATE Compute the per-dimension mean $\mu$ and standard deviation $\sigma$ of $\{\hat{o}_{i+1}\}_{i=1}^{N_{\mathrm{syn}}}$ in $\mathcal{D}_{\mathrm{syn}}$

\STATE Normalize $\{\hat{o}_{i+1}\}_{i=1}^{N_{\mathrm{syn}}}$ using $\mu$ and $\sigma$, and clip the normalized results to a finite range, obtaining $\{\tilde{o}_{i+1}\}_{i=1}^{N_{\mathrm{syn}}}$

\STATE Replace $\{\hat{o}_{i+1}\}_{i=1}^{N_{\mathrm{syn}}}$ in $\mathcal{D}_{\mathrm{syn}}$ with $\{\tilde{o}_{i+1}\}_{i=1}^{N_{\mathrm{syn}}}$

\STATE Initialize the state-action encoder $f_\theta$ and decoder $d_\phi$

\WHILE{Dynamics-prior pretraining}
    \STATE Sample a batch of $(\tilde{o}_t,{a}_t,\tilde{o}_{t+1}) \sim \mathcal{D}_{\mathrm{syn}}$
    \STATE Compute $z_t=f_{\theta}(\tilde{o}_t,{a}_t)$ and $\mathcal{L}_{\mathrm{Aux}}=\|d_{\phi}(z_t)-\tilde{o}_{t+1}\|_2^2$
    \STATE Update $\theta$ and $\phi$ by minimizing $\mathcal{L}_{\mathrm{Aux}}$
\ENDWHILE
\STATE Sample fresh state-action encoder parameters $\theta_0$ from the original random initialization distribution

\STATE Obtain pretrained parameters $\bar{\theta}\gets\theta$ and initialize the state-action encoder with
$
\theta_{\mathrm{init}}=\alpha\bar{\theta}+(1-\alpha)\theta_0
$
\STATE Run the base actor-critic algorithm for online RL with the state-action encoder initialized with $\theta_{\mathrm{init}}$
\end{algorithmic}
\end{algorithm}

\section{Background on AnonMethod and NormMethod}
\label{app:anonmethod_normmethod}

This section introduces AnonMethod and NormMethod, which serve as the backbone RL algorithm and observation normalization module in our experiments. The source code for AnonMethod and NormMethod is included in the supplementary material.

\subsection{NormMethod}
\label{app:normmethod}

NormMethod~\cite{anonymous2026preliminary} is an observation normalization method designed for online RL with low-dimensional observations. Its motivation is that different observation dimensions can have substantially different value ranges. When an auxiliary dynamics prediction loss is applied directly in the raw observation space, dimensions with larger ranges may dominate the prediction loss and its gradients, causing the learned representation to underemphasize dimensions with smaller ranges.

NormMethod normalizes observations before they are used by the value network and the auxiliary next-observation prediction task. Unlike standard normalization schemes that compute statistics over all historical observations, NormMethod is tailored to online RL, where the observation distribution can shift as the policy changes during training. 
It maintains adaptive running statistics from episode-level observation statistics using exponential moving averages, 
and uses these statistics to map raw observations into a normalized observation space with bounded values.
To make the statistics robust to noisy exploration, 
NormMethod excludes episode-level statistics associated with abnormally low returns from the EMA updates. The benefit of NormMethod is not limited to addressing the bottleneck of observation-predictive RL. 
When combined with the model-free method TD3 \cite{DBLP:conf/icml/FujimotoHM18}, NormMethod can also substantially improve its sample efficiency and asymptotic performance. 

\subsection{AnonMethod}
\label{app:anonmethod}

AnonMethod~\cite{anonymous2026preliminary} is a model-free actor-critic method that augments value learning with observation-predictive representation learning. 
In state-based environments, the observation is first normalized by NormMethod. 
The normalized observation and the action are then fed into the state-action encoder of the value network to 
produce a latent state-action representation. 
This representation is used for value prediction and for the auxiliary task of predicting the next \textit{normalized} observation. During online training, the state-action encoder is jointly optimized by the value-learning objective and the auxiliary next-observation prediction objective.

\section{Additional Details for Experimental Setup}

\subsection{Computing Infrastructure}
All experiments were conducted on an x86\_64 server running Ubuntu 24.04.4 LTS (Noble Numbat). The server was equipped with two Intel Xeon Platinum 8368Q CPUs (2.60 GHz; 76 physical cores in total), 251 GiB of system memory, and four NVIDIA GeForce RTX 4090 GPUs. The software environment consisted of Python 3.9.23 and PyTorch 2.6.0.

\subsection{Randomness Control}
For each method or variant on each task, we conducted 5 independent runs using the random seeds $\{42, 99, 123, 520, 668\}$. We seeded the pseudorandom number generators of Python, NumPy, and PyTorch and enabled deterministic execution for cuDNN operations.

\section{Environment Details}
\paragraph{Gym.} This benchmark consists of 5 commonly used locomotion tasks from OpenAI Gym \cite{DBLP:conf/nips/TowersKBCDGKKKP25} in the MuJoCo simulator \cite{DBLP:conf/iros/TodorovET12}, with continuous actions and low-dimensional observations. We use the \texttt{-v5} version. Following AnonMethod, we use the scores of a \textit{deep} variant of TD3 to normalize the scores of other methods when aggregating results.
Deep-TD3 is constructed by enlarging the original TD3 networks so that its number of trainable parameters is comparable to those of AnonMethod and its variants on this benchmark.
Specifically, we increase the depth of the value networks from three to 5 layers, with spectral normalization (SN) \cite{DBLP:conf/icml/GogianuBRCBP21, DBLP:conf/nips/BjorckGW21} applied to the 2nd through 4th layers to stabilize training. Each hidden layer of both the value and actor networks contains 450 units. Deep-SAC, which is built on SAC \cite{DBLP:conf/icml/HaarnojaZAL18}, is constructed following the same principle as Deep-TD3. 
% Building on SAC, it increases the depth of the value networks and also adopts spectral normalization.

\paragraph{DMControl (proprioceptive).} The DeepMind Control Suite (DMControl) \cite{DBLP:journals/corr/abs-1801-00690} is a collection of continuous control tasks built on the MuJoCo simulator. These tasks use low-dimensional proprioceptive data as observations. The maximum total reward for each episode is 1000, making it easy to aggregate results.

\begin{table}[t]
\centering
\begin{tabular}{lccc}
\toprule
\textbf{Environment} & \textbf{Random} & \textbf{Deep-TD3} & \textbf{Deep-SAC} \\
\midrule

\multirow{1}{*}{Ant-v5} 
    & -0.6 & 3908 & 5979 \\

\multirow{1}{*}{Humanoid-v5} 
    & 92.2 & 2897 & 4958 \\

\multirow{1}{*}{HalfCheetah-v5} 
    & -265.0 &  13528 & 14824 \\

\multirow{1}{*}{Hopper-v5} 
    & 25.1 &  3009 & 2926 \\

\multirow{1}{*}{Walker2d-v5} 
    & 5.1 &  4831 & 3640 \\
\bottomrule
\end{tabular}
\caption{Reference scores used to compute the Deep-TD3- and Deep-SAC-normalized results on the Gym benchmark. The Deep-TD3 and Deep-SAC scores are measured at 1M environment steps.}
\label{tab:deepTD3}
\end{table}

\section{Implementation Details}
We use AnonMethod~\cite{anonymous2026preliminary} as the backbone RL algorithm for ProDVI.
Specifically, the state-action encoder is a three-layer multilayer perceptron (MLP) with an Exponential Linear Unit (ELU) activation applied after each layer. SN is applied to the last two layers.
The decoder is a two-layer MLP with ELU activation applied after the first layer.
The value predictor consists of three linear layers.
SN is applied to the first two linear layers, each of which is followed by a ReLU activation.
The policy network is a three-layer MLP where the first two layers are followed by ReLU activations.

Following the design choices of MR.Q \cite{DBLP:conf/iclr/FujimotoD0TR25}, AnonMethod uses the LAP replay buffer \cite{fujimoto2020equivalence} for prioritized sampling during training, and adopts the same target network update strategy, where all target parameters are periodically synchronized with their online counterparts. The value learning loss is computed using the Huber loss \citep{fujimoto2020equivalence}. All networks are trained with the AdamW \citep{DBLP:conf/iclr/LoshchilovH19} optimizer. 

ProDVI adopts the observation normalization method NormMethod, proposed in AnonMethod, during online RL to balance the auxiliary losses across observation dimensions. 
The source code for ProDVI is provided in the supplementary material.

% \subsection{Synthetic Data Generation}
% The synthetic data generation procedure by LLM-generated dynamics priors is detailed in Lines 3–8 of Algorithm~\ref{algorithm:prodvi}.

\begin{table*}[t]
    \centering
    \begin{tabular}{ccccccc}
        \toprule
        \multirow{2}{*}{\textbf{{Environment Steps}}}  & \multirow{2}{*}{\textbf{Metrics}} & 
        \multicolumn{5}{c}{\textbf{Synthetic Transitions}} \\
        \cmidrule(lr){3-7}
        && {{1M}} & {{3M}} & {{5M}} & {7M} & {10M} \\
        \midrule     
        \multirow{2}{*}{0.25M}
        &Mean &
        1.15 & 
        1.13 &
        \textbf{1.24} & 
        1.22 & 
        0.95 \\

        &IQM & 
        0.94 & 
        1.05 &
        \textbf{1.09} & 
        1.04 & 
        0.92 \\
        
        \midrule

        \multirow{2}{*}{0.50M}
        &Mean & 
        1.40 & 
        1.47 &
        1.54 & 
        \textbf{{1.58}} & 
        1.42 \\
        
        &IQM & 
        1.20 & 
        1.23 &
        1.29 & 
        \textbf{{1.33}} & 
        {1.24} \\
        
        \midrule    

        \multirow{2}{*}{1.00M}
        &Mean & 
        1.71 &  
        {1.69} &
        1.69 & 
        \textbf{{1.76}} & 
        {1.69} \\

        &IQM & 
        1.44 & 
        {{1.45}} &
        1.45 & 
        \textbf{{1.49}} & 
        {1.43} \\
      
        \bottomrule
    \end{tabular}
    \caption{
    Sensitivity study on the number of synthetic transitions used for dynamics-prior pretraining on Gym. We report aggregated Deep-TD3-normalized scores of AnonMethod + ProDVI under different synthetic dataset sizes, with $\alpha$ fixed to $1.0$. Bold numbers indicate the best performance under the same metric and environment-step budget.
    }
    \label{hyper_table1}
\end{table*}

\begin{table*}[t]
    \centering
    \begin{tabular}{cccccc}
        \toprule
        \multirow{2}{*}{\textbf{{Environment Steps}}}  & \multirow{2}{*}{\textbf{Metrics}} & 
        \multicolumn{4}{c}{\textbf{Interpolation Coefficient $\alpha$}} \\
        \cmidrule(lr){3-6}
        
        && 0.2& 0.5 & 0.8 & 1.0 \\
        \midrule     
        \multirow{2}{*}{0.25M}
        &Mean &
        1.12 & 
        \textbf{1.33} &
        {1.24} & 
        1.22 \\

        &IQM & 
        0.97 & 
        \textbf{1.15} &
        0.99 & 
        1.04 \\
        
        \midrule

        \multirow{2}{*}{0.50M}
        &Mean & 
        1.45 & 
        \textbf{1.64} &
        1.58 & 
        1.58 \\
        
        &IQM & 
        1.10 & 
        \textbf{1.40} &
        1.25 & 
        {1.33} \\
        
        \midrule    

        \multirow{2}{*}{1.00M}
        &Mean & 
        1.74 &  
        \textbf{1.85} &
        \textbf{1.85} & 
        {1.76} \\

        &IQM & 
        1.45 & 
        \textbf{1.54} &
        1.53 & 
        {1.49} \\
      
        \bottomrule
    \end{tabular}
    \caption{
    Sensitivity study on the interpolation coefficient $\alpha$ on Gym. We report aggregated Deep-TD3-normalized scores of AnonMethod + ProDVI under different values of $\alpha$, using 7M synthetic transitions for dynamics-prior pretraining. Bold numbers indicate the best performance under the same metric and environment-step budget.
    }
    \label{hyper_table2}
\end{table*}

\subsection{Hyperparameters}
\paragraph{AnonMethod and NormMethod.} The default hyperparameters of AnonMethod and NormMethod in state-based settings are summarized in Table~\ref{tab:hyperparameters}.  

\begin{table*}[t]
\centering

\begin{tabular}{lll}
\toprule
\textbf{Components} & \textbf{Hyperparameter} & \textbf{Value} \\
\midrule

\multirow{4}{*}{Value Learning} 
    & n-step returns & 1 \\
    & Auxiliary loss weight $\lambda_\text{Aux}$ & \makecell[l]{Gym: 10 \\  DMControl: 2} \\
    & Discount factor $\gamma$ & 0.99 \\

\midrule

\multirow{4}{*}{NormMethod} 
    & Queue size of history episode $|\mathcal{Q}|$& \makecell[l]{Gym: 200 \\ DMControl: 400} \\
    & EMA coefficient $\beta$ & \makecell[l]{Gym: $1-1 / 200$ \\ DMControl: $1-1 / 400$} \\
    & Clip bounds $(-O, O)$ & (-10, 10) \\
\midrule

\multirow{2}{*}{TD3} 
    & Target policy noise $\sigma$ & $ \mathcal{N}(0,0.2^2)$ \\
    & Target policy noise clipping $c$ & (-0.5, 0.5) \\
\midrule

\multirow{2}{*}{LAP} 
    & Priority exponent & 0.4 \\
    & Minimum priority & 1.0 \\
\midrule
\multirow{7}{*}{Optimization} 
    & Optimizer & AdamW \\
    & Learning rate & 3e-4 \\
    & Weight decay & 1e-4 \\
    & Mini-batch size & 256 \\
    & Target update frequency & 250 \\
    & Gradient updates per training step &  \makecell[l]{value network: 1 \\ policy network: 0.5} \\

\midrule

\multirow{2}{*}{Exploration} 
    & Initial random exploration time steps & 25k \\
    & Exploration noise & $ \mathcal{N}(0,0.1^2)$ \\

\midrule

\multirow{1}{*}{Observation Encoder} 
    & Structure & identity function \\
\midrule

\multirow{4}{*}{State-Action Encoder} 
    & Hidden dim & 450 \\
    & $z_{sa}$ dim & 450 \\
    & Activation function & ELU \\
    & Gradient clip norm & 20 \\
\midrule

\multirow{3}{*}{Long-term value predictor} 
    & Hidden dim & 450 \\
    & Activation function & ReLU \\
    & Gradient clip norm & 20 \\
\midrule

\multirow{4}{*}{Decoder} 
    & Structure & MLP \\
    & Hidden dim & 450 \\
    & Activation function & ELU \\
    & Gradient clip norm & 20 \\
\midrule

\multirow{2}{*}{Policy Network} 
    & Hidden dim & 450 \\
    & Activation function & ReLU \\
\bottomrule
\end{tabular}
\caption{Default hyperparameters of AnonMethod and NormMethod.}
\label{tab:hyperparameters}
\end{table*}

\begin{table*}[t]
\centering
\begin{tabular}{lll}
\toprule
\textbf{Components} & \textbf{Hyperparameter} & \textbf{Value} \\
\midrule

\multirow{2}{*}{Dynamics Prior Distillation} 
    & Number of synthetic transitions & 7M \\
    & Interpolation coefficient $\alpha$ & 0.5 \\

\midrule

\multirow{1}{*}{Observation Normalization} 
    & Clip bounds $(-O, O)$ & (-10, 10) \\
\midrule

\multirow{4}{*}{Pretraining} 
    & Optimizer & AdamW \\
    & Learning rate & 3e-4 \\
    & Weight decay & 1e-4 \\
    & Mini-batch size & 256 \\

\bottomrule
\end{tabular}
\caption{Default hyperparameters of ProDVI.}
\label{tab:hyperparameters2}
\end{table*}

\paragraph{ProDVI.}
For the underlying RL algorithms and the observation normalization method NormMethod, we keep all hyperparameters the same as those used by the corresponding methods. 
The key hyperparameters introduced by ProDVI are the number of synthetic transitions used for dynamics-prior pretraining and the interpolation coefficient $\alpha$. 

We study the effects of these two hyperparameters on AnonMethod + ProDVI using the Gym benchmark. Table~\ref{hyper_table1} summarizes the benchmark-level aggregated performance under different numbers of synthetic transitions, with $\alpha$ fixed to $1.0$. For all settings in this study, the number of pretraining epochs and the batch size are fixed to $5$ and $256$, respectively. The results show that performance generally improves as the number of synthetic transitions increases, but starts to degrade when too many synthetic transitions are used. We hypothesize that using too few synthetic transitions may prevent the dynamics priors from being sufficiently distilled into the state-action encoder. On the other hand, using too many synthetic transitions may overtrain the encoder on the synthetic dynamics-prior dataset, making it harder to adapt during subsequent online RL with real environment interactions, which is consistent with the phenomenon of plasticity loss \cite{DBLP:conf/collas/AbbasZM0M23, DBLP:journals/nature/DohareHLRMS24}. Since $7$M synthetic transitions achieves the best results in the largest number of aggregated metrics, we use $7$M as the default number of synthetic transitions in ProDVI.
Based on this setting, we further investigate the effect of the interpolation coefficient $\alpha$. As shown in Table~\ref{hyper_table2}, $\alpha=0.5$ achieves the best overall performance on the Gym benchmark. Therefore, we set $\alpha=0.5$ as the default value for ProDVI. Note that this procedure is not intended to be an exhaustive search over all possible hyperparameter combinations. Nevertheless, the simple selection strategy described above yields a default configuration that performs well on both Gym and DMControl.

\section{Prompt Templates}
This section presents the prompt templates used by ProDVI. ProDVI uses a code-based interface to communicate environment information to the LLM. The core prompting pipeline consists of three stages. First, we construct a structured Python-class description of the target environment. Second, we prompt the LLM to produce a textual analysis of the environment dynamics. Third, we prompt the LLM to convert the textual dynamics analysis into executable Python code that maps a current observation-action pair to an approximate next observation.

\subsection{Environment Descriptions}
The first step of ProDVI is to construct a structured description of the target environment. Similar to Eureka~\cite{DBLP:conf/iclr/MaLWHBJZFA24}, we represent the environment description as a Python class. This code-based format provides a structured interface for subsequent prompts, allowing the LLM to reason about the observation space, action space, and available environment configuration parameters when generating dynamics analyses and executable dynamics-prior code.

\paragraph{Docstrings.}
The docstring of the Python class contains a brief description of the environment, together with the dimension-wise semantics of the observation and action spaces. For Gym tasks, this information can be obtained directly from the official environment documentation\footnote{\url{https://gymnasium.farama.org/environments/mujoco/}}.  For example, the Ant task documentation provides a description of the environment as well as the meanings of the observation and action dimensions. For DMControl tasks, the high-level task description is available from the DMControl documentation \cite{DBLP:journals/corr/abs-1801-00690}.
The dimension-wise semantics of proprioceptive observations and actions, however, are not provided in the same explicit format as Gym documentation. Therefore, we prompt an LLM to generate a Python program for automatically interpreting the observation and action spaces of DMControl tasks. The generated program uses MuJoCo model metadata to assign physically meaningful labels to observation dimensions, including joint states, body states, and sensor measurements. It also maps each action dimension to the corresponding named actuator and retrieves its admissible control range.
The prompt template used to generate this parser is shown in Prompt~\ref{lst:dmcontrol_parser_prompt}.

\begin{listing}[tb] 
\caption{Prompt template for generating a parser of DMControl observation and action spaces.} 
\label{lst:dmcontrol_parser_prompt} 
\begin{lstlisting}[style=promptstyle] 
You are an expert in reinforcement learning and the DMControl benchmark with proprioceptive observations. We would like to inspect the semantic meaning of each dimension in the observation and action spaces of the {target domain} environment. Please provide code that satisfies this requirement. 
\end{lstlisting} 
\end{listing}

\paragraph{Internal class variables.}
In addition to qualitative information about the observation and action spaces, ProDVI can also use quantitative environment information when such information is available. We store this information as internal variables of the Python class. For example, both Gym and DMControl environments are defined using XML configuration files, which specify the robot morphology, body structure, actuator definitions, joint properties, and physical parameters such as body sizes and masses.

To extract such quantitative information, we prompt the LLM with the Python class containing the docstring and the corresponding XML configuration file. The LLM is then asked to parse the XML content and store the extracted information as internal variables in the \texttt{\_\_init\_\_} method of the environment class. The prompt template is shown in Prompt~\ref{lst:xml}. 
For environments beyond Gym and DMControl, a similar environment-description construction procedure can be applied as long as metadata about the target environment is available from technical documentation, manuals, or other materials.
\begin{listing*}[tb] 
\caption{Prompt template for extracting quantitative environment configuration parameters from XML files and storing them as internal class variables.} 
\label{lst:xml} 
\begin{lstlisting}[style=promptstyle] 
You are an expert in physics, robotics, and reinforcement learning.

The following is a partial program for simulating {target environment}.
The docstring of the {TargetEnvironment} class provides information about
the observation and action spaces of the environment:
```
import math
import numpy as np

class {TargetEnvironment}:
    """
    Environment Description
    -----------------------
    {Environment Description}
    
    Action Space
    ------------
    {Action Space}
    
    Observation Space
    -----------------
    {Observation Space}
    """

    def __init__(self):
        """
        Initializes environmental configs and robotic states.
        """ 

```
In addition, we have the XML configuration file of this robot:
```
{XML file content}
```
Now, please parse the XML file and write the extracted information into
the `__init__` method of the {TargetEnvironment} class.
\end{lstlisting} 
\end{listing*}
The resulting class description combines semantic information from the docstring with quantitative configuration parameters stored as internal variables. This class is used as the environment description in subsequent prompts.

\subsection{Dynamics Analysis Template}
After obtaining the structured description of the target environment, ProDVI prompts the LLM to analyze the environment dynamics. The goal of this stage is not to obtain executable code directly, but to first elicit a structured textual explanation of how actions affect observations. This intermediate analysis encourages the LLM to decompose the dynamics into interpretable sub-processes before code generation. The dynamics analysis prompt is shown in Prompts~\ref{lst:analysisA} and~\ref{lst:analysisB}.

\begin{listing*}[tb] 
\caption{Prompt template for generating a textual dynamics analysis (Part A). Part A provides the structured environment description and the interface of the \texttt{do\_simulation} function.} 
\label{lst:analysisA} 
\begin{lstlisting}[style=promptstyle] 
You are an expert in physics, robotics, and reinforcement learning.

We are developing a program for simulating a robot. The following is a
partial implementation of the robot environment ({target environment}),
where the do_simulation function remains to be implemented:
```
import math
import numpy as np

class {TargetEnvironment}:
    """
    {docstrings}
    """

    def __init__(self):
        """
        Initializes environmental configs and robotic states.
        """
        {internal variables}
    
    def do_simulation(self, observation: np.ndarray,
                      action: np.ndarray) -> np.ndarray:
        """
        This function applies the provided actions to the current 
        observations and returns the updated observations.
    
        Parameters
        ----------
        observation : np.ndarray
            The current observation of the environment.
            See the class docstring for its detailed structure.
    
        action : np.ndarray
            The current action applied to the environment.
            See the class docstring for its detailed structure.
    
        Returns
        -------
        next_observation : np.ndarray
            The updated observation after one simulation step.
        """
        # TODO: Implement simulation stepping logic here
        pass

```
The `__init__` method of {TargetEnvironment} defines the relevant configuration parameters of the robot.
\end{lstlisting} 
\end{listing*}

\begin{listing*}[tb] 
\caption{Prompt template for generating a textual dynamics analysis (Part B). Part B specifies the required format for decomposing how actions and current observations affect next observations.} 
\label{lst:analysisB} 
\begin{lstlisting}[style=promptstyle] 
Based on the information provided in {TargetEnvironment} and your prior
knowledge, please analyze how the observation space is affected by the
action space. In other words, explain how the current observation and the
input action may determine the next observation.

Please follow these requirements:
-1 The process by which actions affect observations should be decomposed into several logically organized steps.
-2 The analysis should cover as many observation dimensions as possible.

For each step, please use the following format:
<example>
Step x: <title>
Source variables: <variables that cause the effect>
Target variables: <variables affected by the source variables>
Principle: <how the source variables affect the target variables>
Sub-process: <a detailed description of the mechanism by which the source variables affect the target variables>
</example>
\end{lstlisting} 
\end{listing*}

The output of this prompt is a textual dynamics analysis. It provides a structured decomposition of the approximate causal and physical relationships between actions, current observations, and next observations. This analysis is then used as additional context for generating executable dynamics-prior code.

\subsection{Code Generation Template}
Given the environment description and the textual dynamics analysis, ProDVI further prompts the LLM to instantiate the analyzed dynamics into executable Python code. The generated code implements the \texttt{do\_simulation} function, which takes the current observation and action as inputs and returns an approximate next observation.

The code generation prompt reuses the dynamics analysis prompt and appends the generated textual analysis as context, which is shown in Prompt~\ref{lst:code}.
\begin{listing*}[tb] 
\caption{Prompt template for generating executable dynamics-prior code from the textual dynamics analysis.} 
\label{lst:code} 
\begin{lstlisting}[style=promptstyle] 
{Dynamics Analysis Template}

The following is the dynamics analysis for this environment:
```
{dynamics analysis}
```
Now, based on the analysis above, please instantiate each analysis step as one or more executable Python functions. Then combine these functions into a complete implementation of do_simulation, which maps the current observation and current action to the next observation.

The final output should be valid Python code that completes the `do_simulation` function.

\end{lstlisting} 
\end{listing*}
The resulting program is used as the LLM-generated dynamics-prior function in ProDVI. We emphasize that this function is not used as a faithful simulator and is not used for planning. Instead, it is executed on randomly sampled observation-action pairs to generate synthetic transitions, which are then used to pretrain the state-action encoder of the value network.

\section{Algorithm Comparisons}
Table~\ref{tab:algos} summarizes the main algorithmic differences among TD3, SAC, AnonMethod, and their variants considered in our experiments. Here, $\text{Loss}_{Q}$ denotes the loss function used for value learning, while \textit{Backbone RL Algorithm} indicates the underlying actor-critic algorithm on which each method is built. \textit{Deeper Networks} indicates whether a method uses networks deeper than its corresponding vanilla backbone.
Deep-TD3 and Deep-SAC introduce only minimal modifications to the original algorithms. 
In contrast, AnonMethod and its SAC-based variant incorporate additional techniques for stable value learning and auxiliary representation learning. The ablated variants, AnonMethod w/o Aux and AnonMethod-SAC w/o Aux, remove this auxiliary task while retaining the other components.

\begin{table*}[t]
\small
\setlength{\tabcolsep}{1mm}
\centering

\begin{tabular}{lcccc|cccc}
\toprule
\textbf{Components} & \textbf{TD3} & \textbf{Deep-TD3} & \textbf{SAC} & \textbf{Deep-SAC} & \textbf{\makecell{AnonMethod \\ w/o Aux}} & \textbf{\makecell{AnonMethod-SAC \\ w/o Aux}}  & \textbf{AnonMethod} & \textbf{AnonMethod-SAC} \\
\midrule
{$\text{Loss}_{Q}$} 
     & \multicolumn{4}{c|}{MSE} & \multicolumn{4}{c}{Huber Loss} \\
{Optimizer} 
     & \multicolumn{4}{c|}{Adam} & \multicolumn{4}{c}{AdamW} \\
{Observation Normalization} 
     & \multicolumn{4}{c|}{$\times$} & \multicolumn{4}{c}{\checkmark} \\
{Gradient Clipping} 
     & \multicolumn{4}{c|}{$\times$} & \multicolumn{4}{c}{\checkmark} \\
{Prioritized Replay Buffer} 
     & \multicolumn{4}{c|}{$\times$} & \multicolumn{4}{c}{\checkmark} \\
{Deeper Networks} 
     & $\times$ & \checkmark & $\times$ & \checkmark & \multicolumn{4}{c}{\checkmark} \\
{Spectral Normalization} 
     & $\times$ & \checkmark & $\times$ & \checkmark & \multicolumn{4}{c}{\checkmark} \\
{Auxiliary Task} 
     & \multicolumn{4}{c|}{$\times$} & $\times$ & $\times$ & \checkmark & \checkmark \\
{Backbone RL Algorithm} 
     & TD3 & TD3 & SAC & SAC & TD3 & SAC & TD3 & SAC \\
\bottomrule
\end{tabular}
\caption{Comparison of TD3, SAC, AnonMethod, and their variants used in our experiments.}
\label{tab:algos}
\end{table*}

\section{Additional Details of Experiments}
This section provides additional experimental details for the dynamics prediction experiments in Section 5.4. The dynamics predictor is trained using the mean squared error (MSE) loss. We use the AdamW optimizer with a learning rate of $3\times10^{-4}$, a weight decay of $1\times10^{-4}$, and a minibatch size of 256. 
The transition data used for training are collected from the replay buffers of Deep-TD3 agents. Both the input observations and target next observations are normalized using the mean and standard deviation computed over the entire replay buffer, and are then clipped to the range $[-10, 10]$.

% Unlike standard supervised learning settings, data collected in reinforcement learning is subject to distributional shifts induced by the evolving policy. Therefore, we do not construct a separate held-out evaluation set or emphasize held-out evaluation loss in this experiment. 
We report the training loss of the dynamics prediction task, which reflects how efficiently different initializations adapt to the non-stationary transition data encountered during training. For each dynamics prediction task on the Gym benchmark, we run 5 independent seeds. At 0.25M and 0.50M gradient updates, we evaluate each dynamics predictor on 10 minibatches sampled from the corresponding replay buffer and report the average MSE across these minibatches as the training loss. We then aggregate the seed-level training losses and report their mean together with 95\% bootstrap confidence intervals obtained by resampling the 5 seeds.

\section{Complete Results}
\begin{figure*}[t]
\centering
\includegraphics[width=0.33\textwidth]{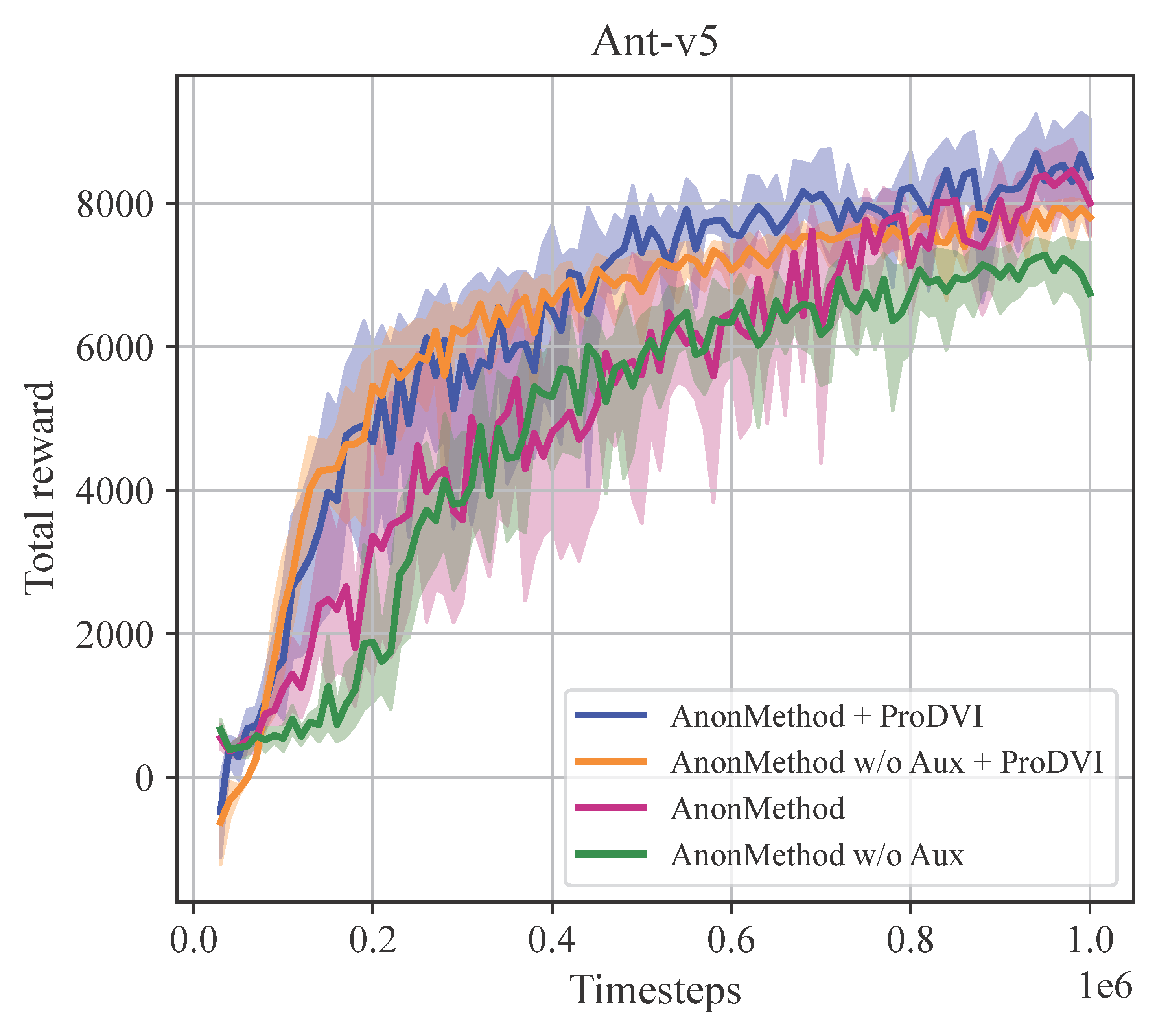}
\includegraphics[width=0.33\textwidth]{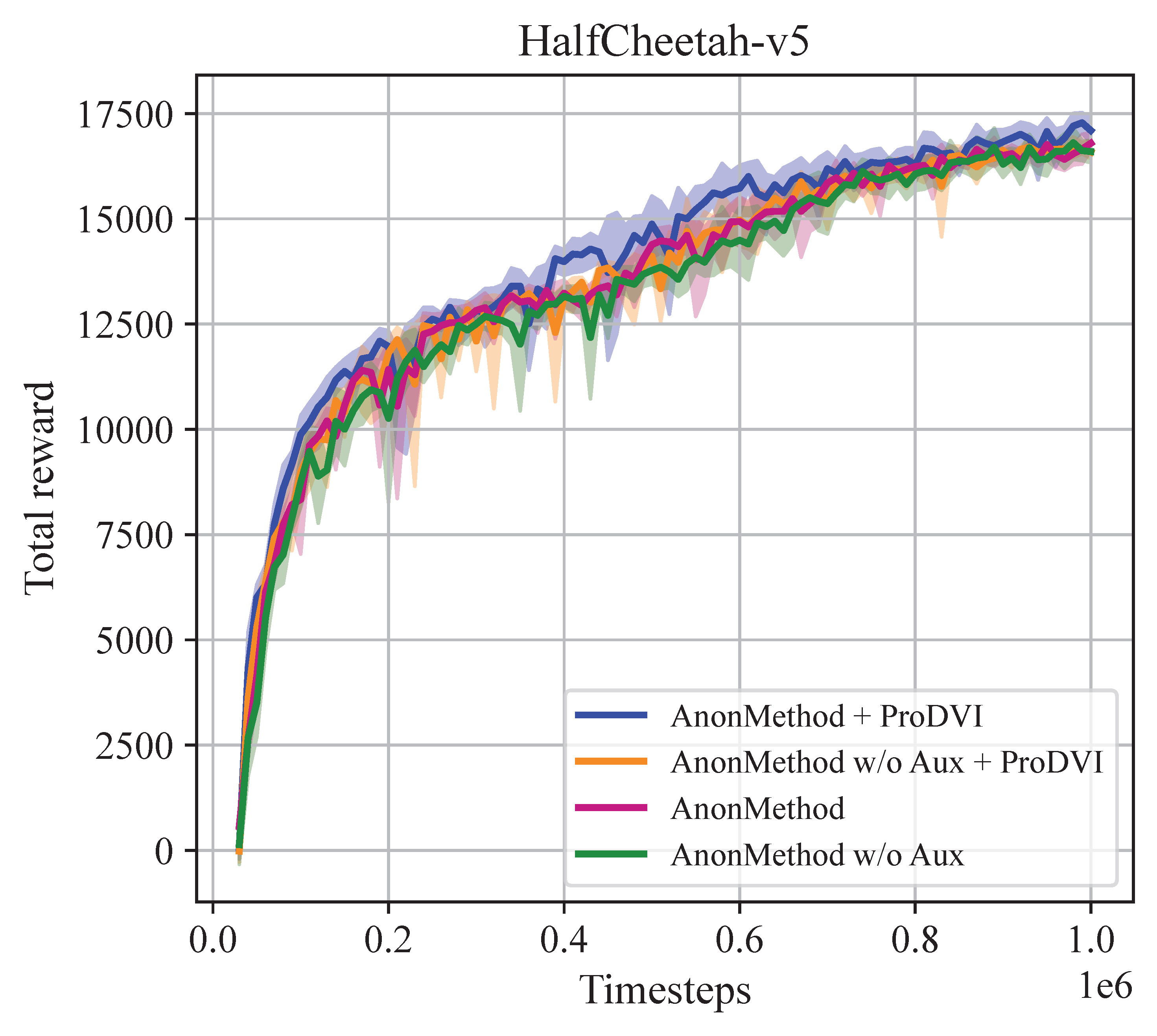}
\includegraphics[width=0.33\textwidth]{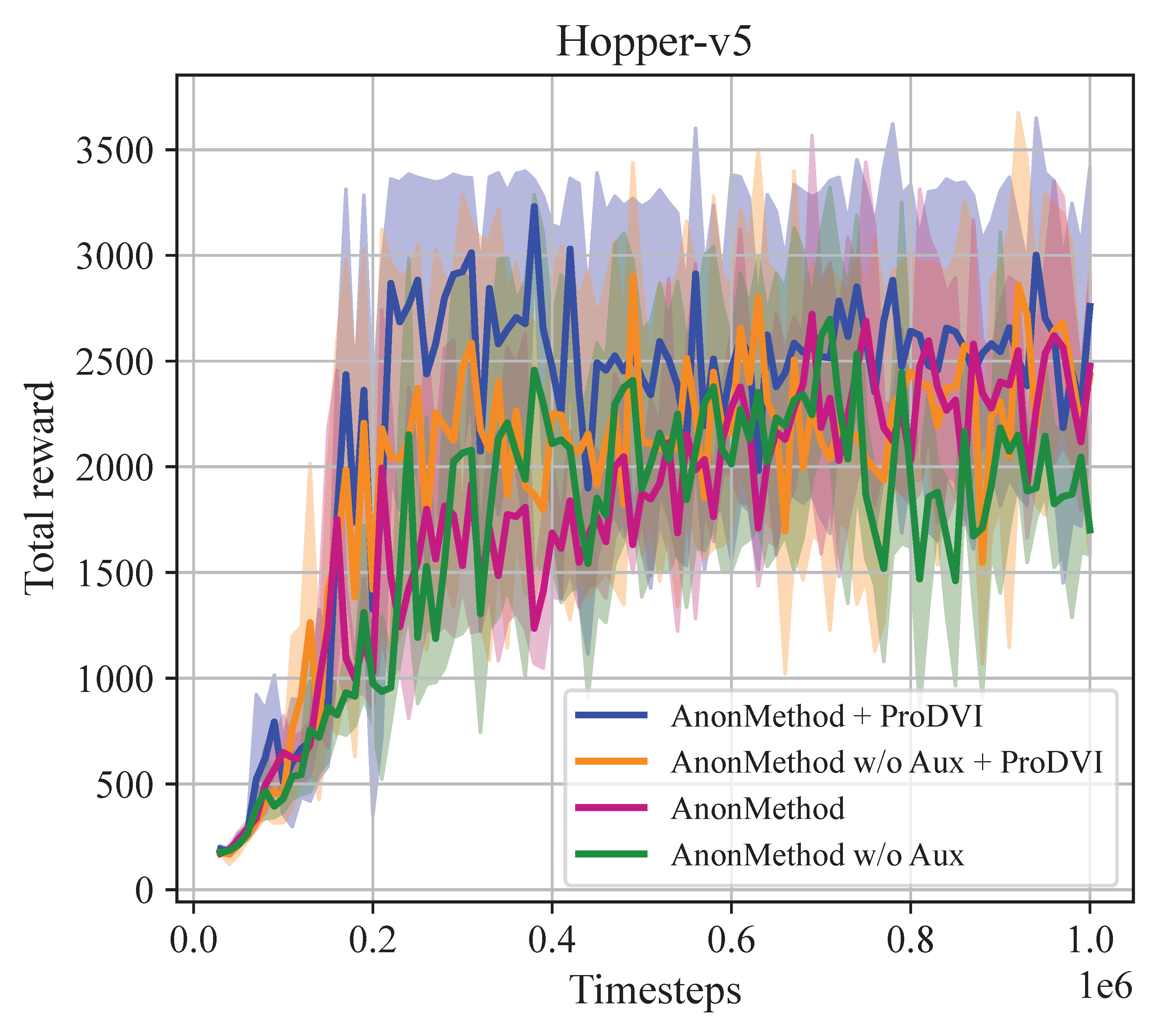}
\includegraphics[width=0.33\textwidth]{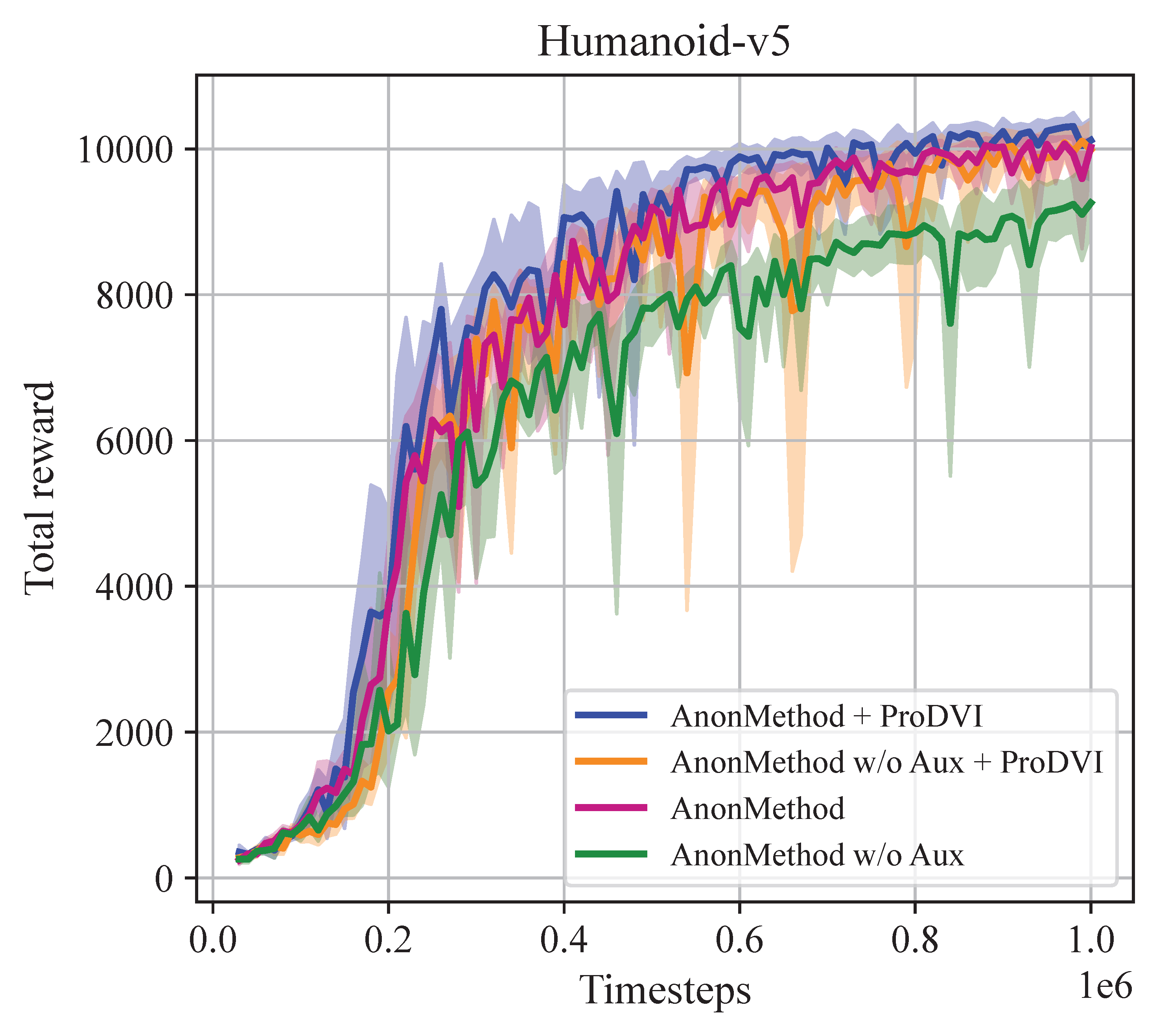}
\includegraphics[width=0.33\textwidth]{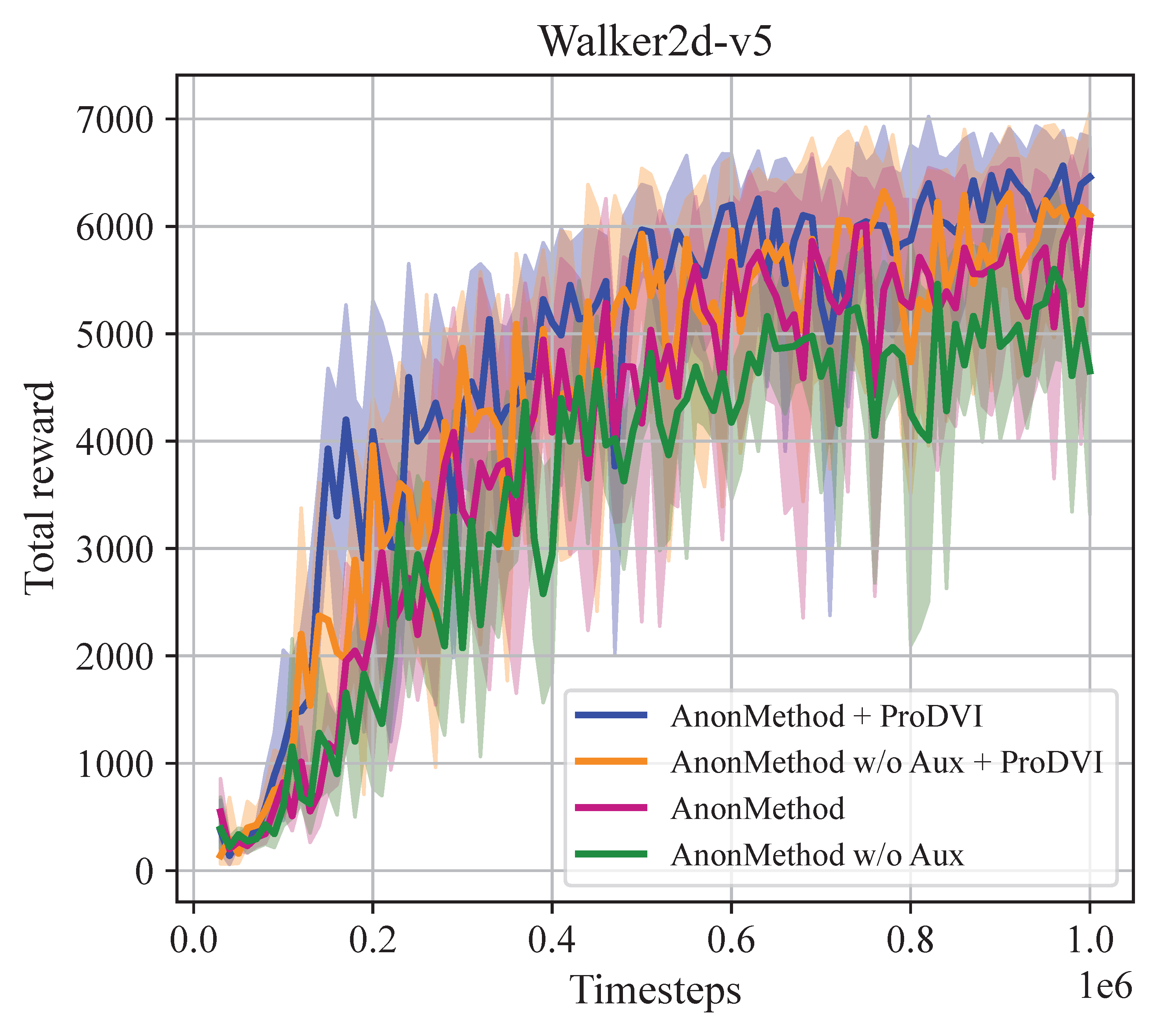}
\includegraphics[width=0.33\textwidth]{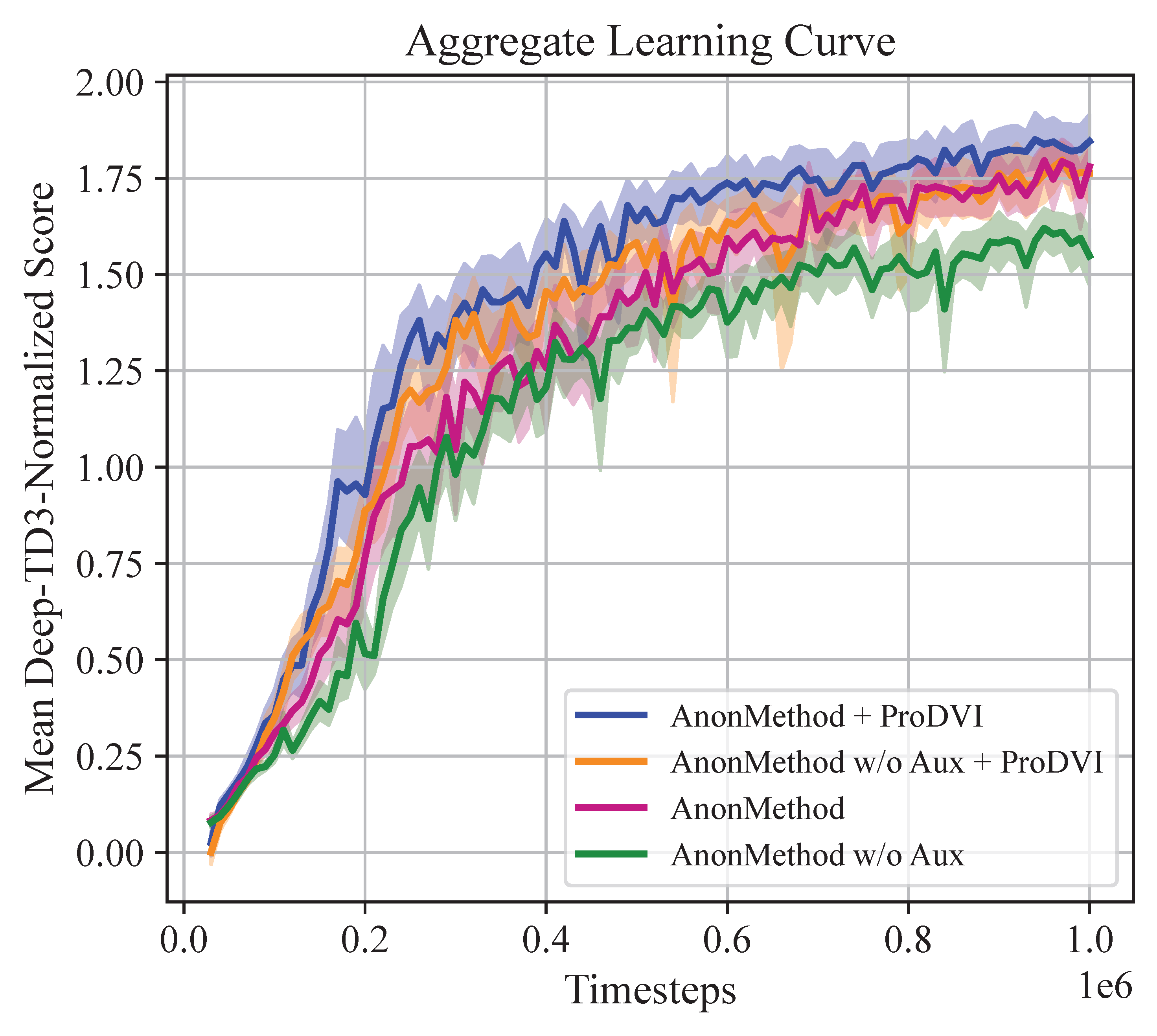}
\caption{
Per-task and aggregate learning curves on the Gym benchmark for AnonMethod w/o Aux, AnonMethod, and their ProDVI-enhanced variants. The aggregate curve reports the mean Deep-TD3-normalized score across the 5 Gym tasks. Shaded area captures a 95\% bootstrap confidence interval.
}
\label{curves1}
\end{figure*}

\begin{figure*}[t]
\centering
\includegraphics[width=0.33\textwidth]{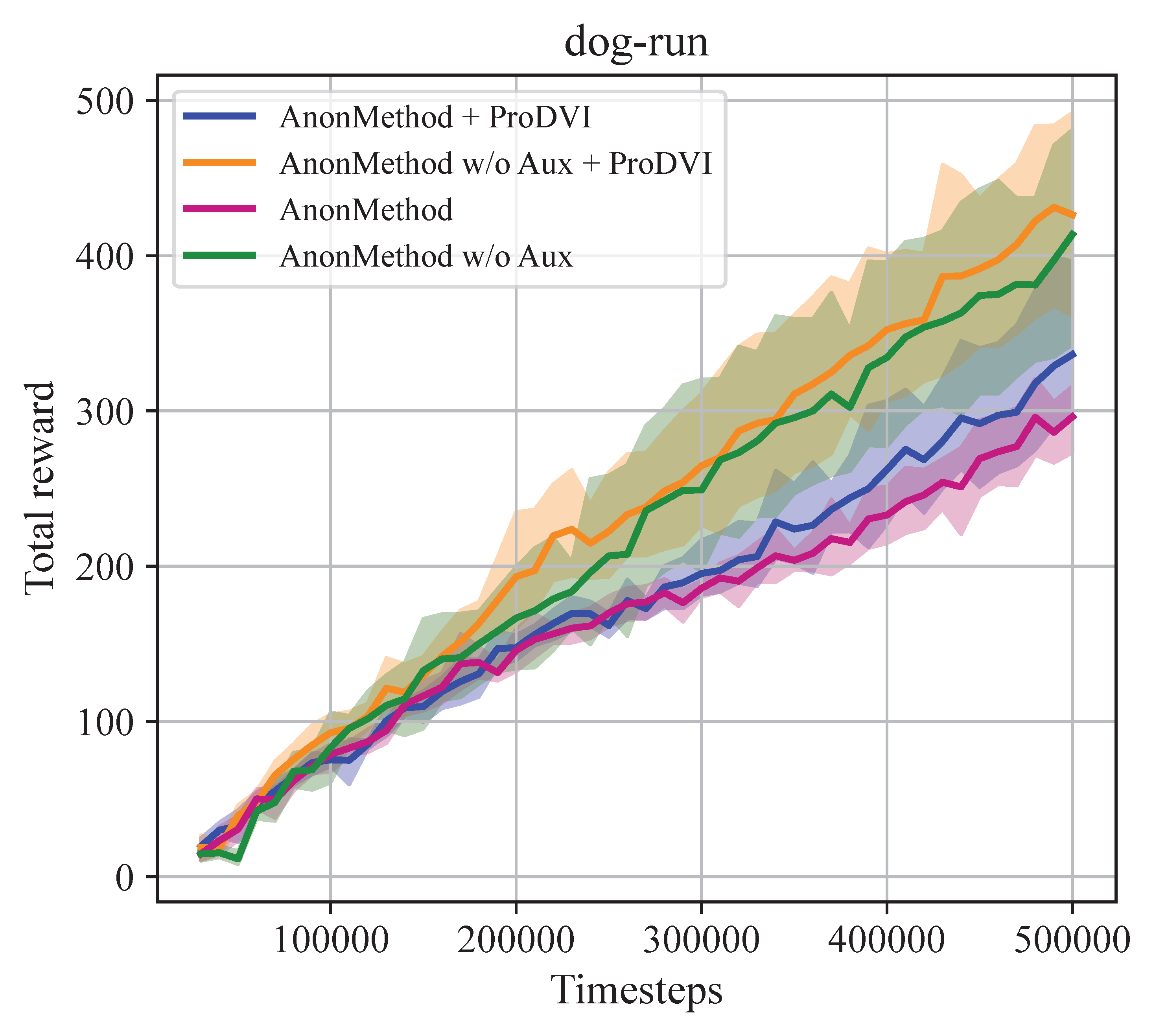}
\includegraphics[width=0.33\textwidth]{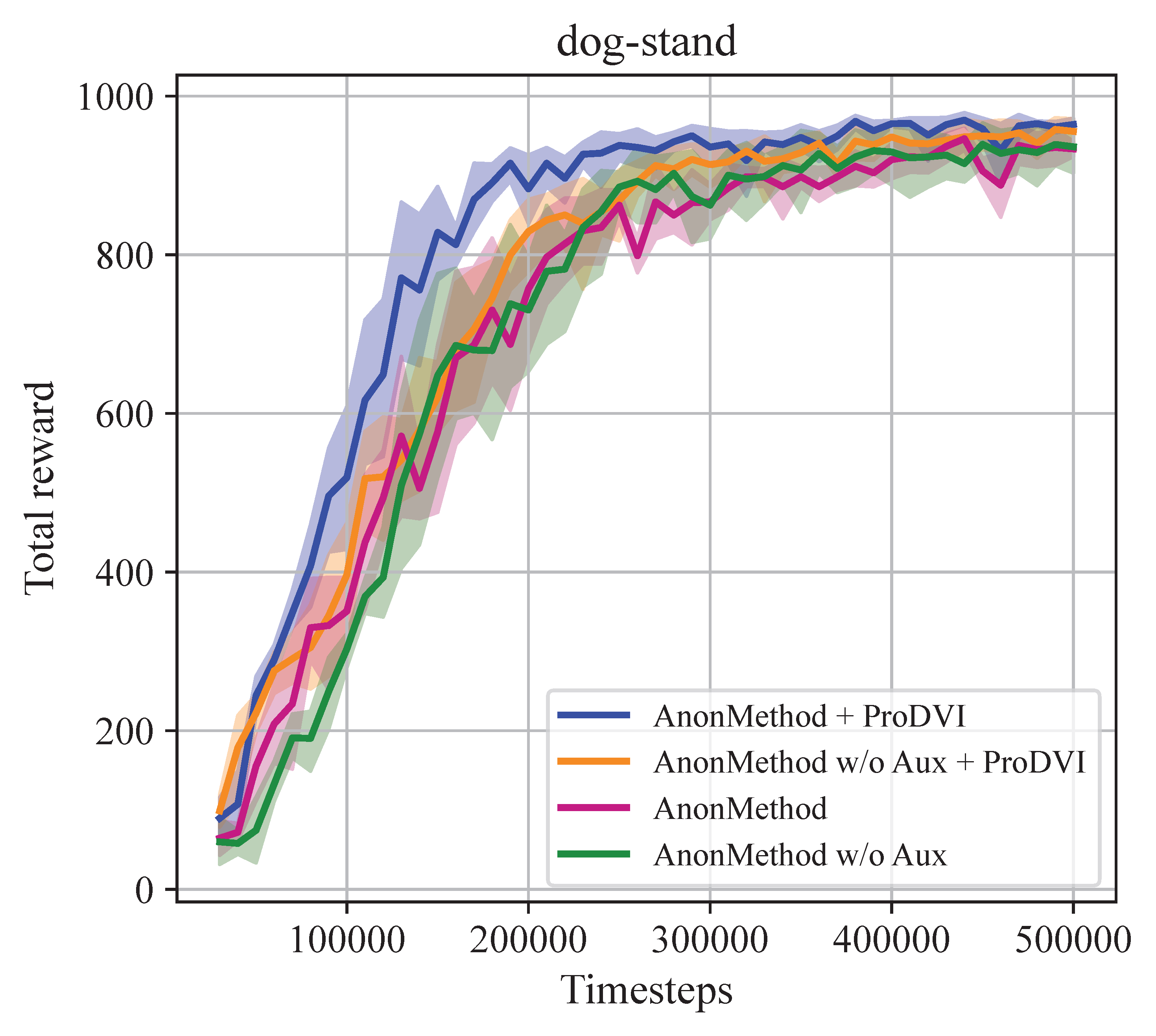}
\includegraphics[width=0.33\textwidth]{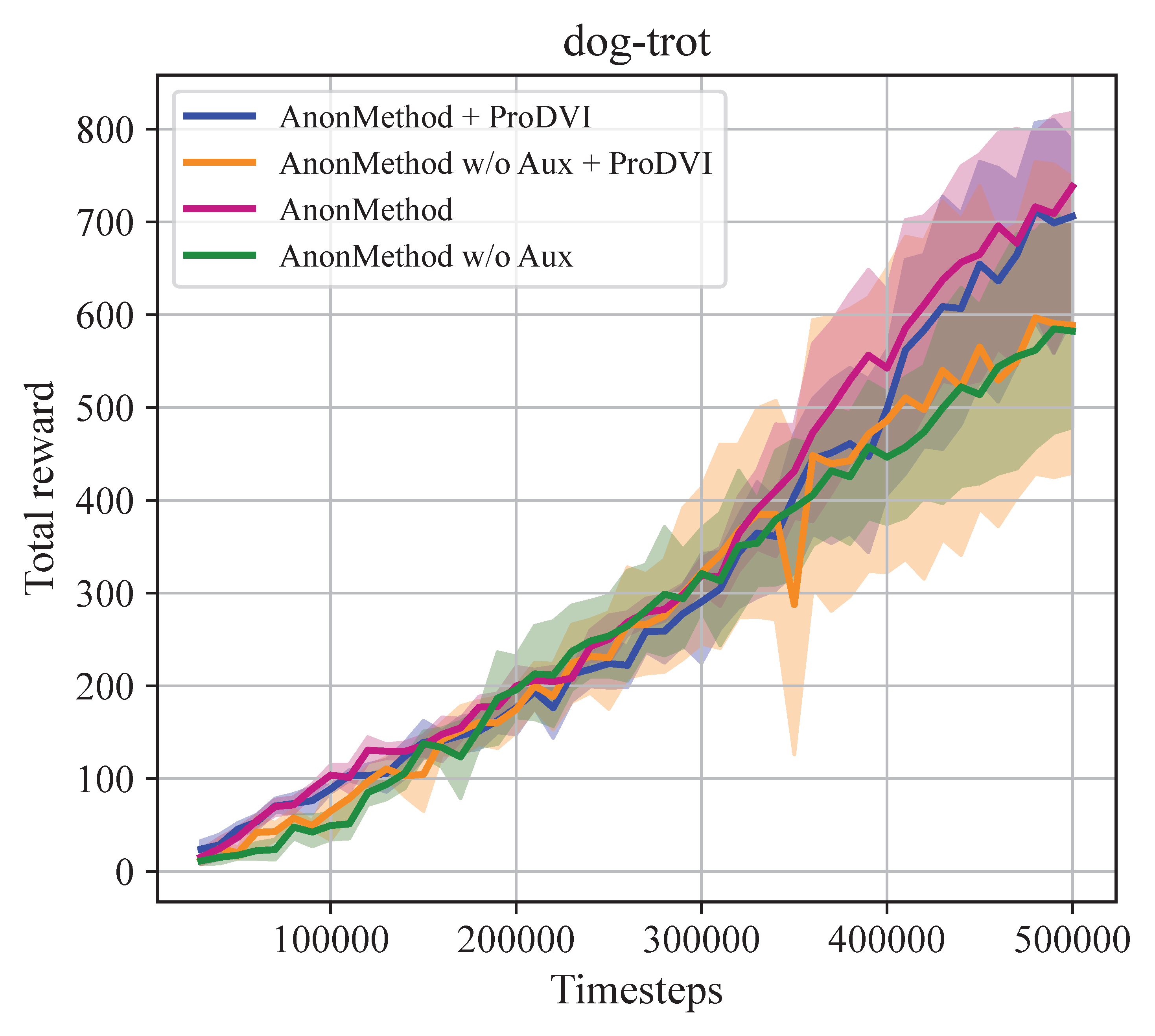}
\includegraphics[width=0.33\textwidth]{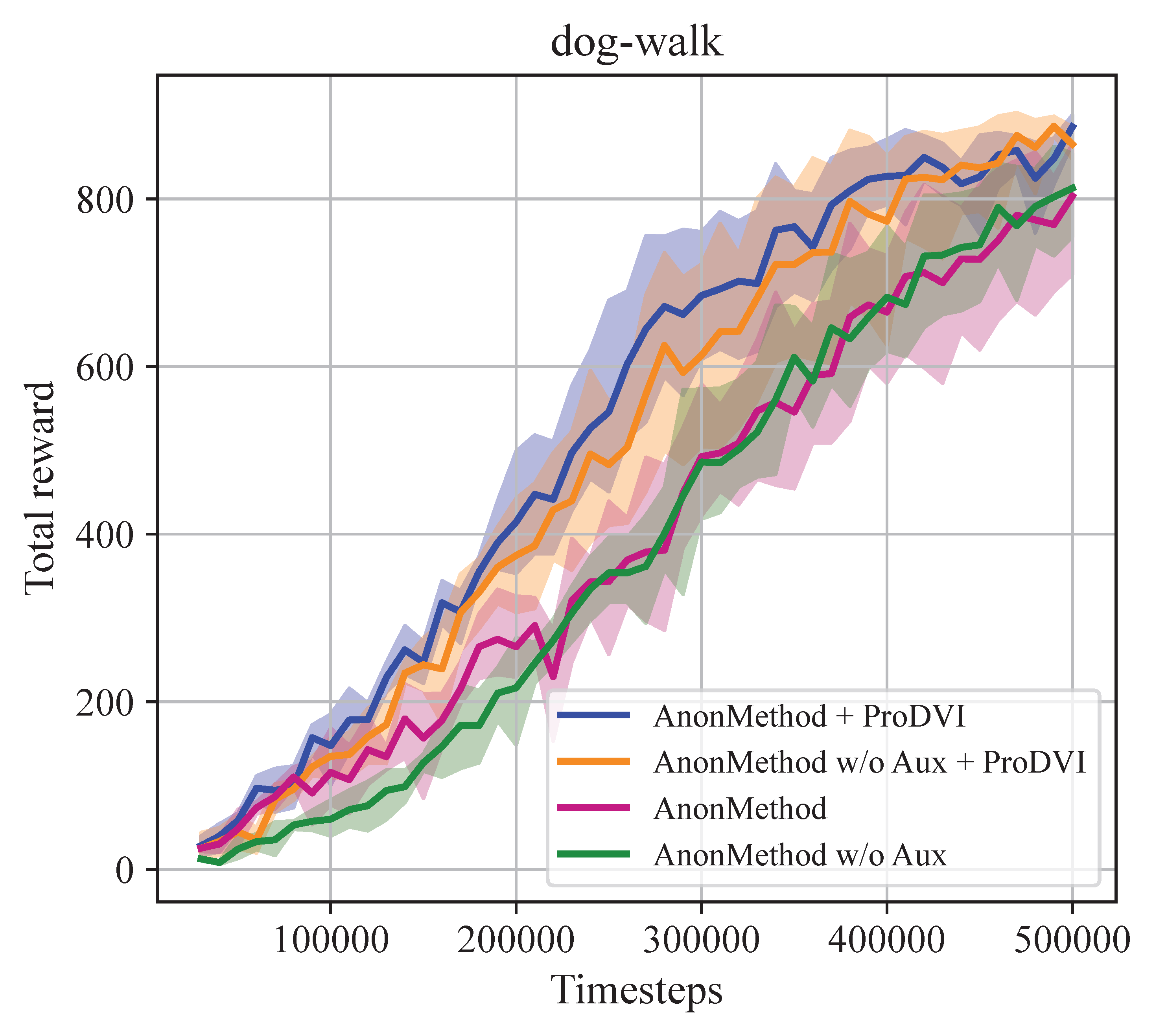}
\includegraphics[width=0.33\textwidth]{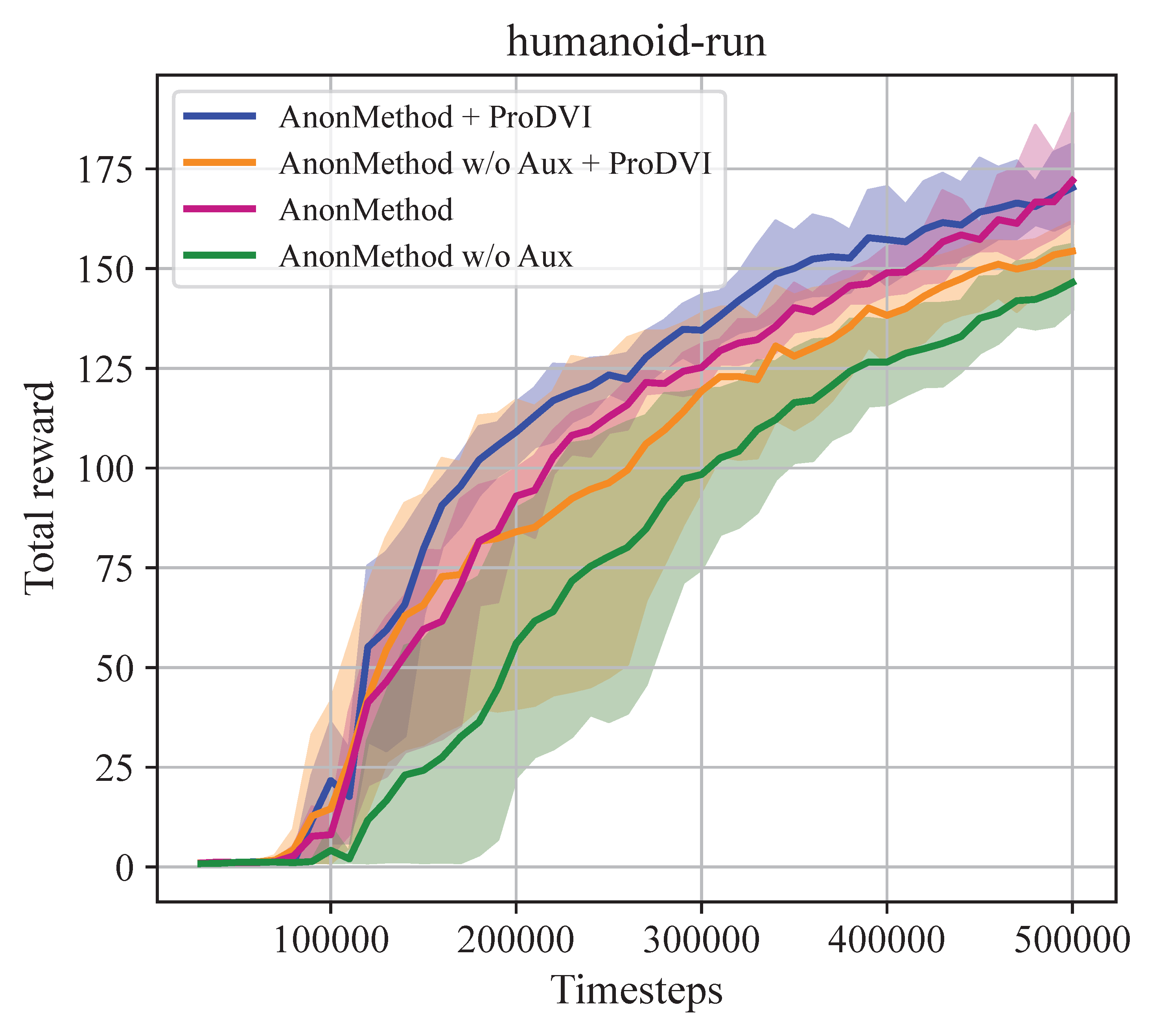}
\includegraphics[width=0.33\textwidth]{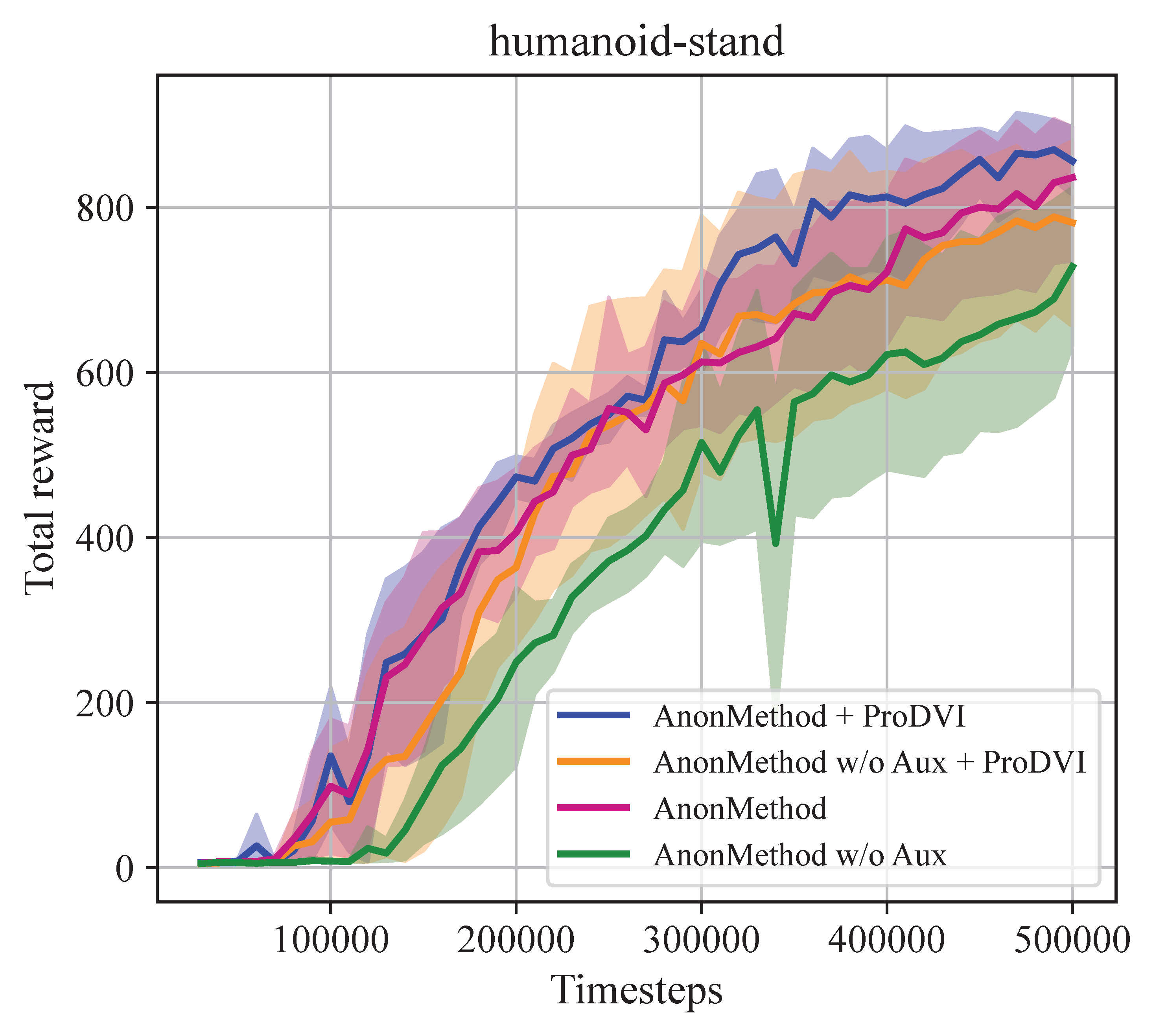}
\includegraphics[width=0.33\textwidth]{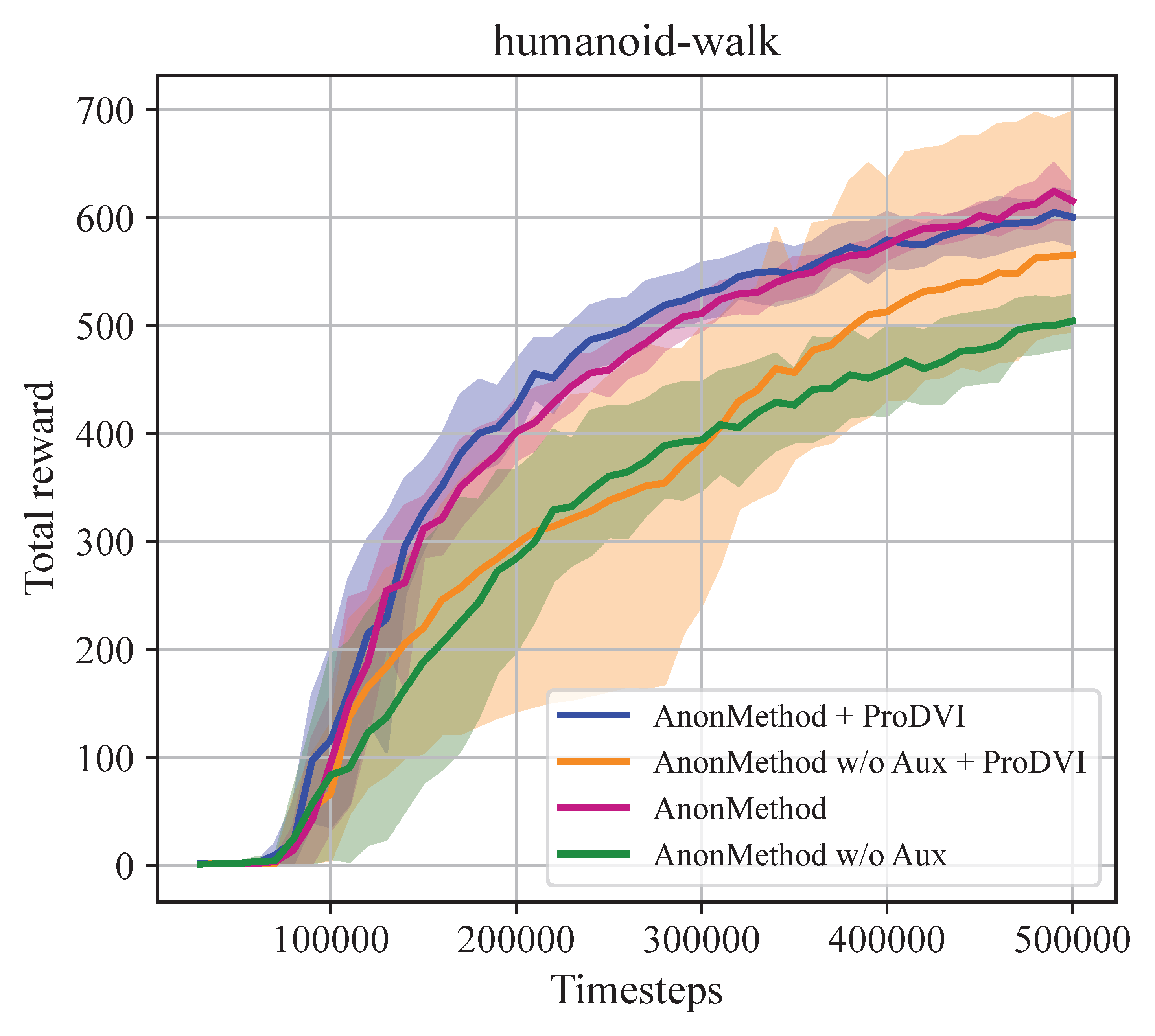}
\includegraphics[width=0.33\textwidth]{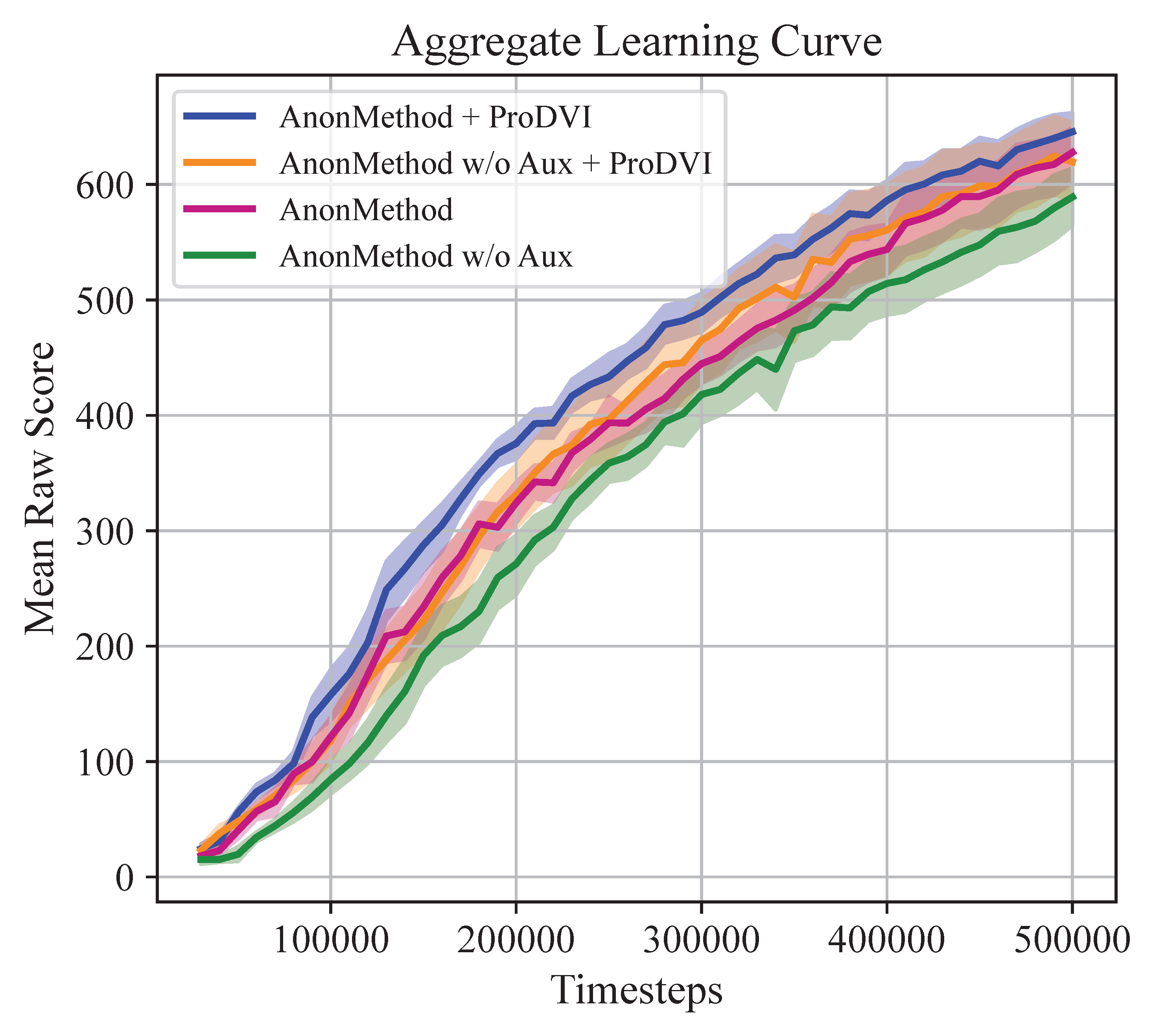}
\caption{Per-task and aggregate learning curves on the DMControl tasks for AnonMethod w/o Aux, AnonMethod, and their ProDVI-enhanced variants. The aggregate curve reports the mean performance across the 7 DMControl tasks. Shaded areas indicate 95\% bootstrap confidence intervals.}
\label{curves2}
\end{figure*}

\begin{figure*}[t]
\centering
\includegraphics[width=0.33\textwidth]{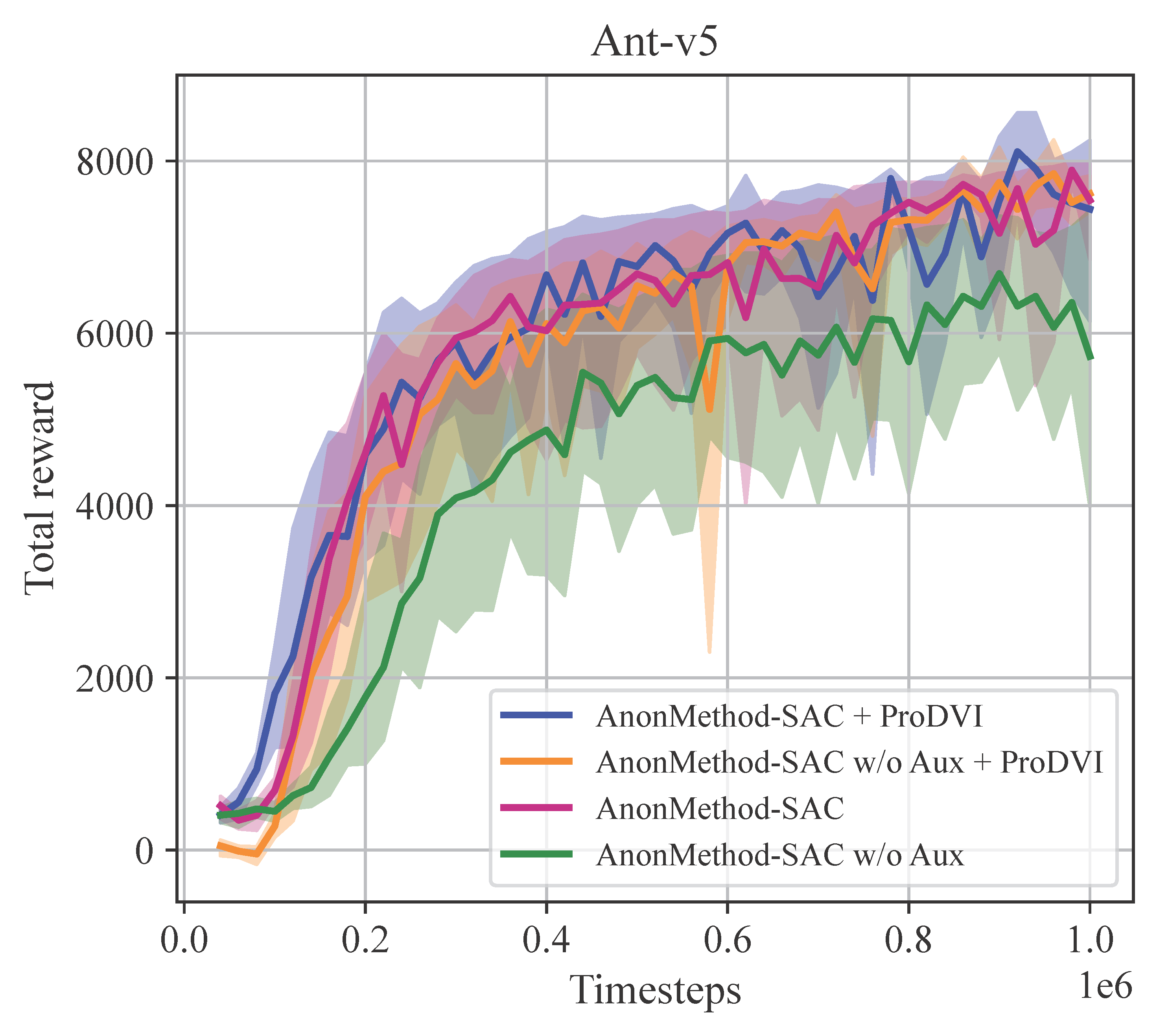}
\includegraphics[width=0.33\textwidth]{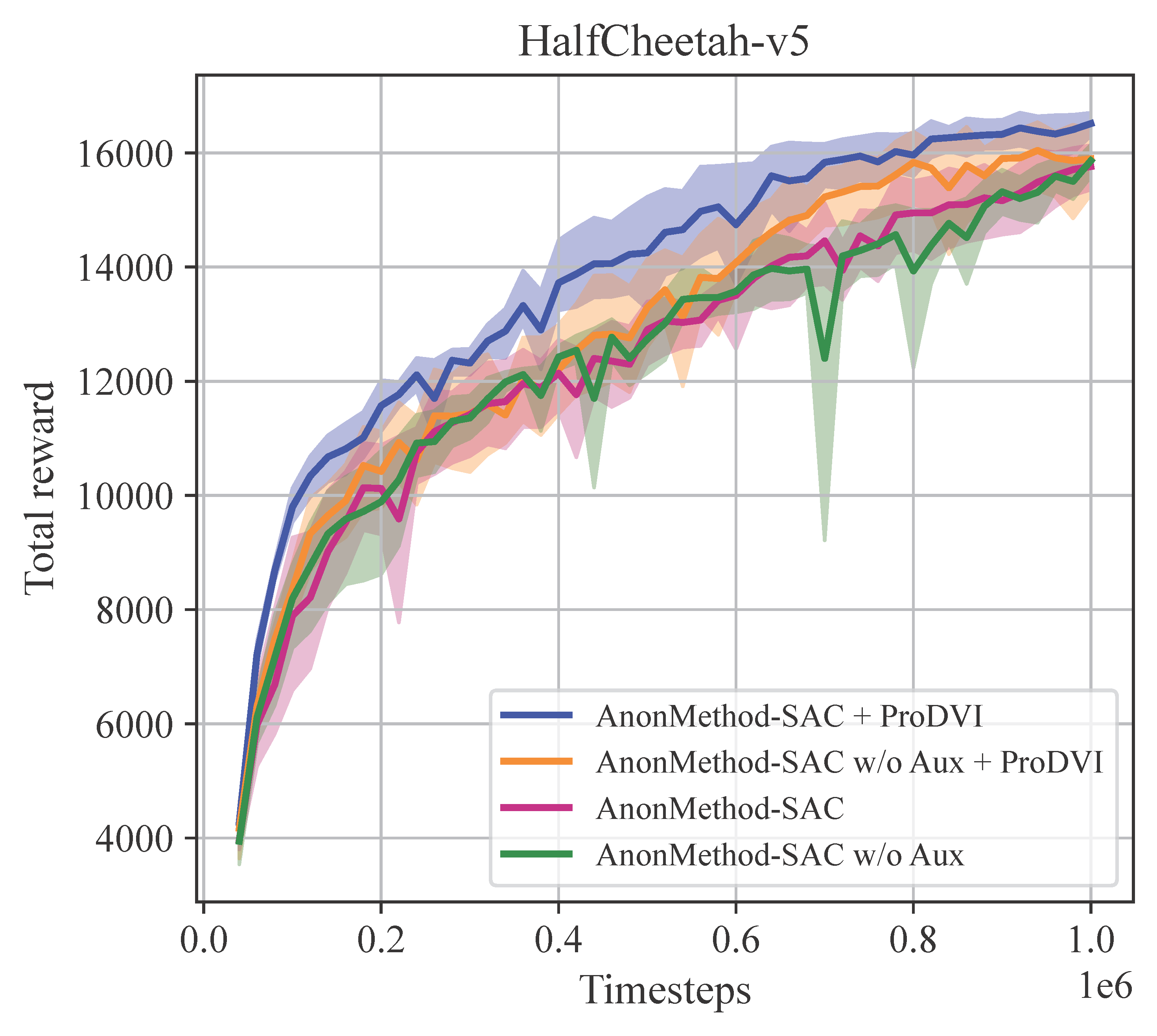}
\includegraphics[width=0.33\textwidth]{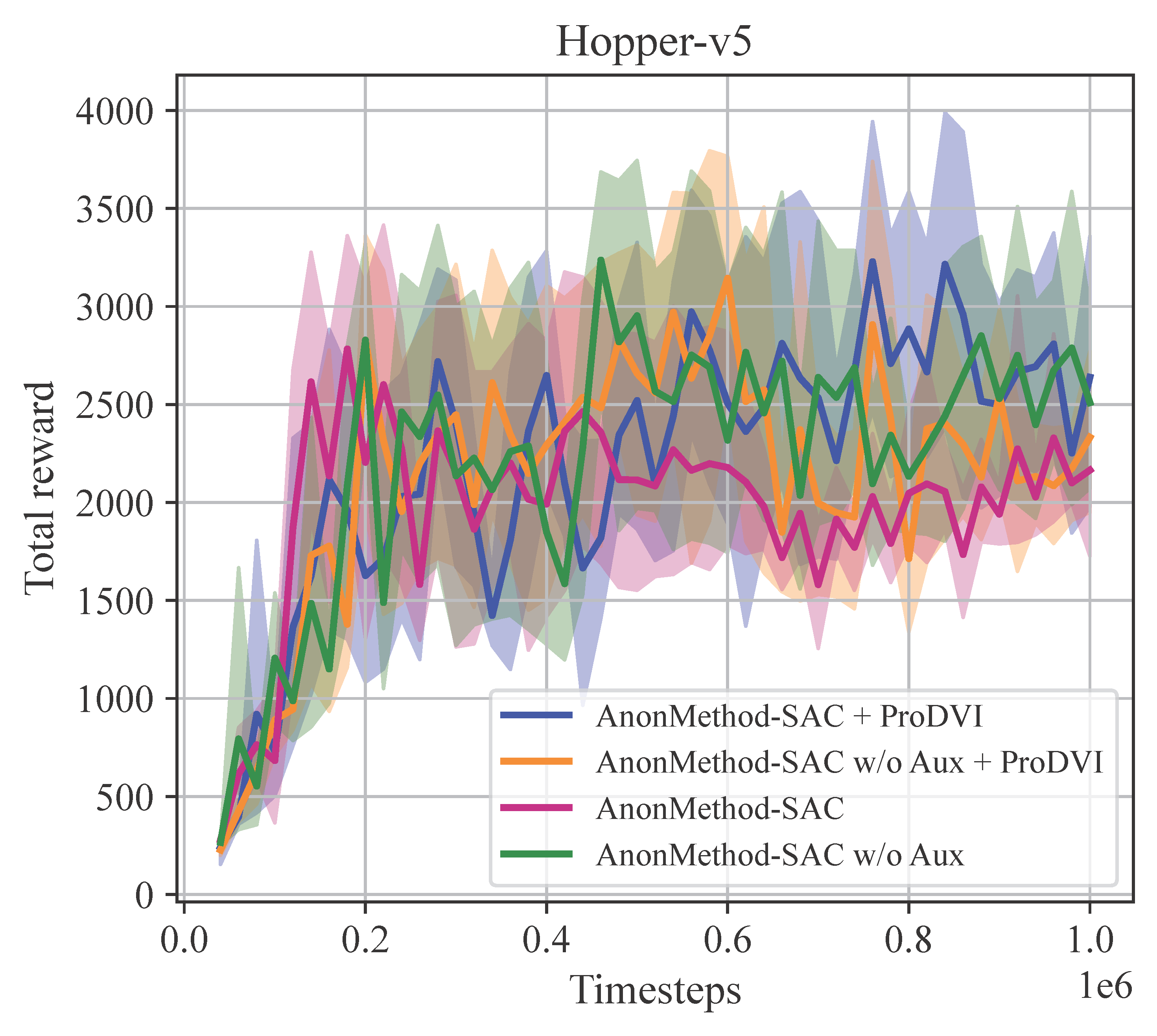}
\includegraphics[width=0.33\textwidth]{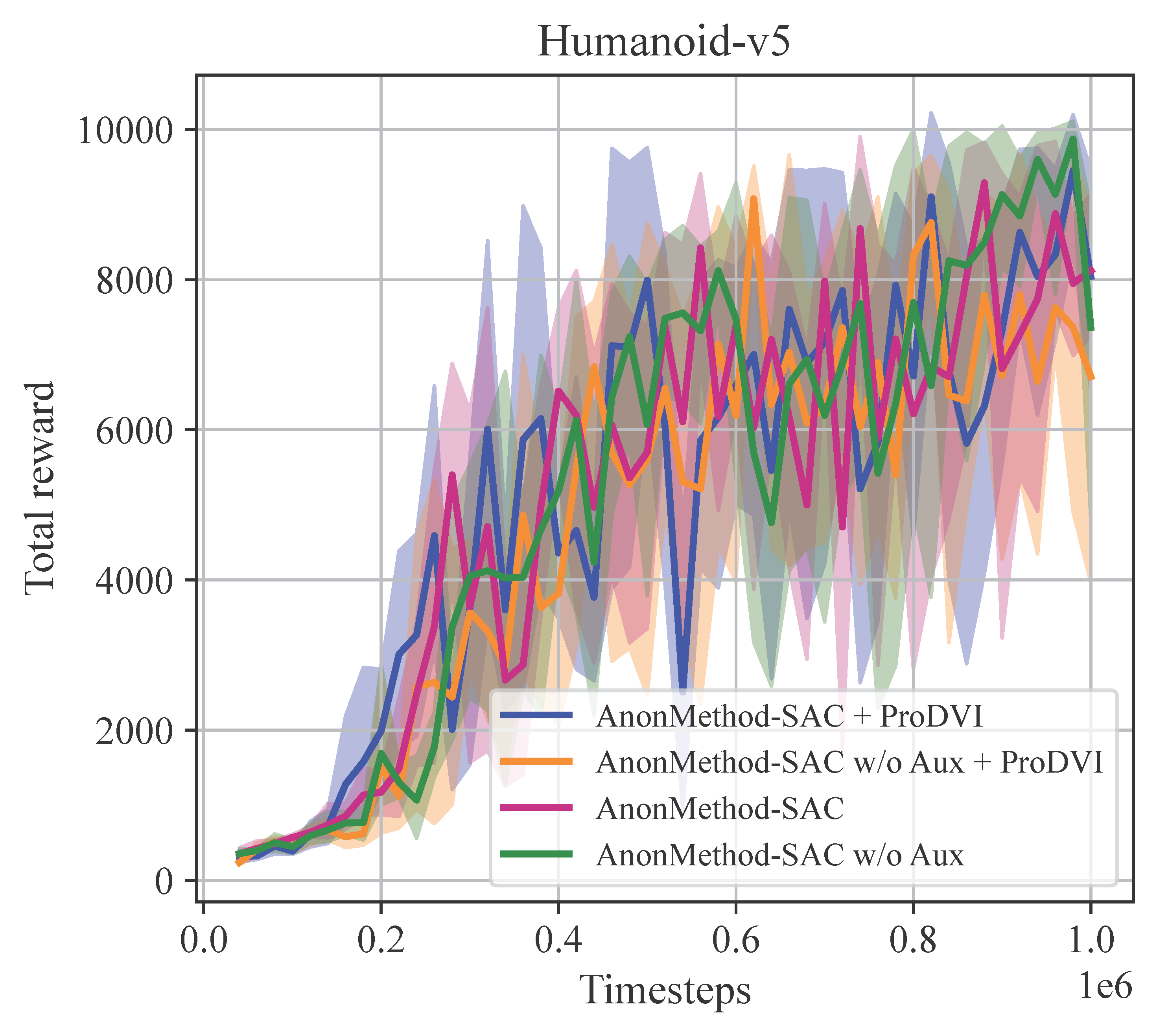}
\includegraphics[width=0.33\textwidth]{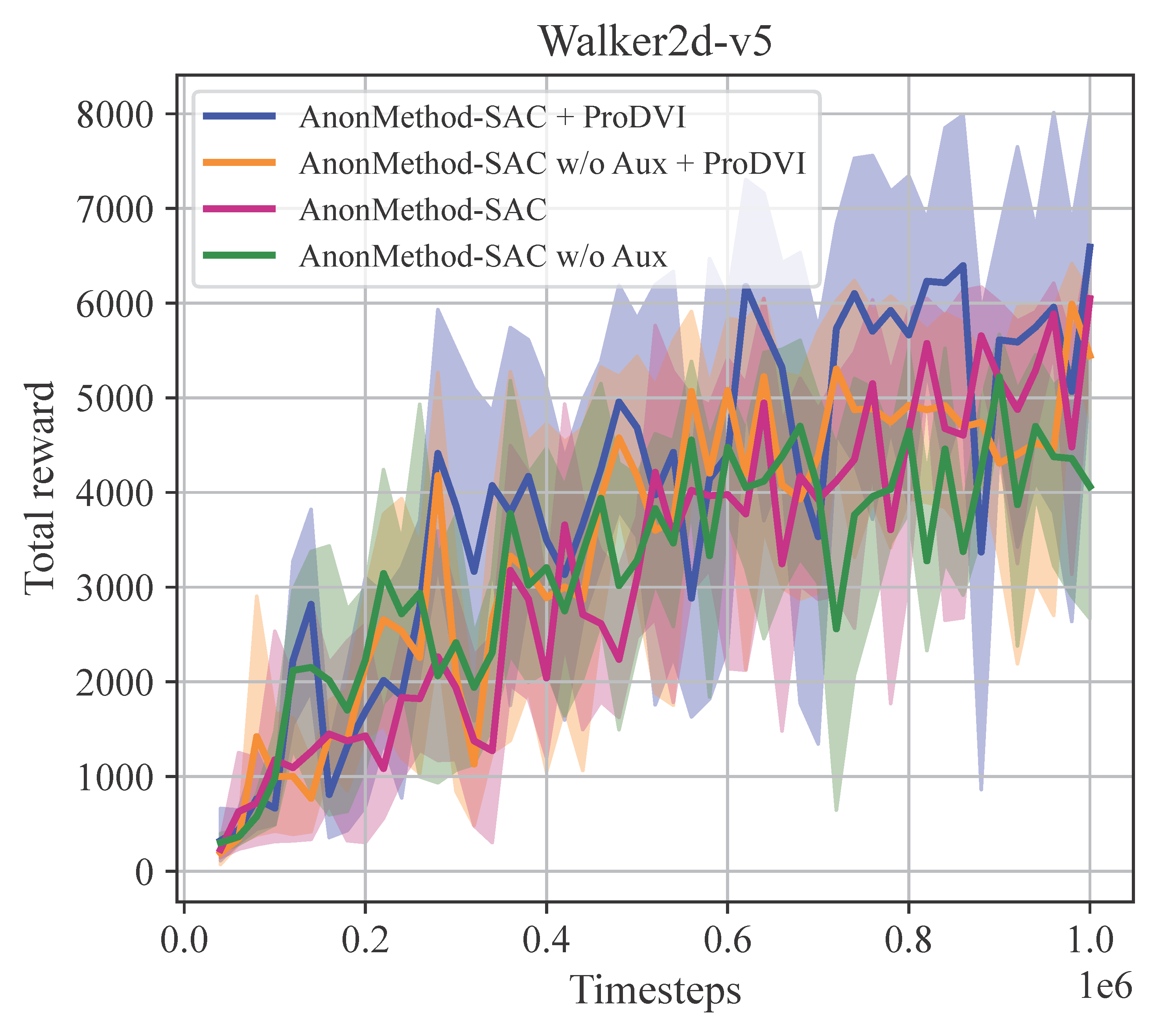}
\includegraphics[width=0.33\textwidth]{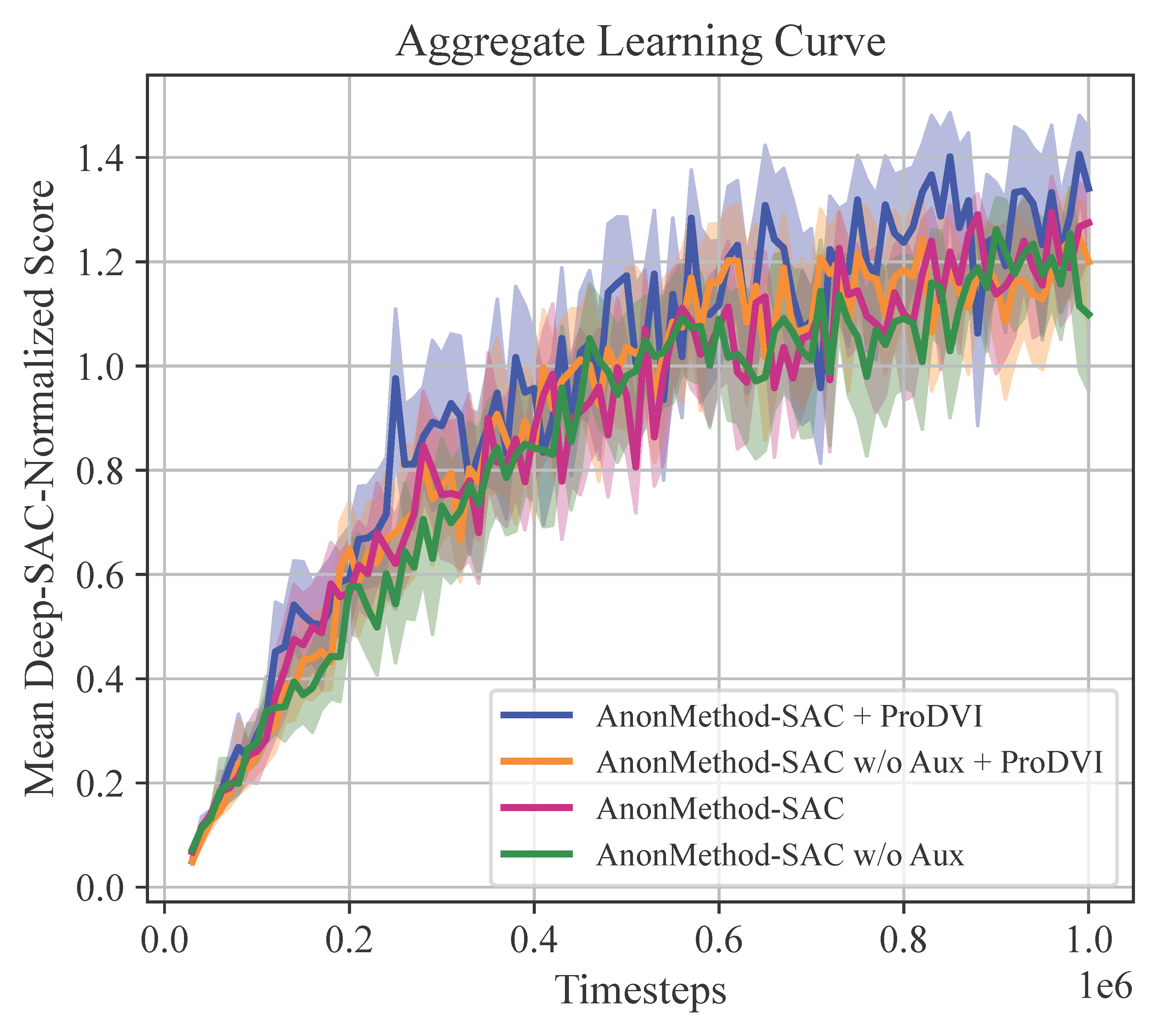}
\caption{
Per-task and aggregate learning curves on the Gym benchmark for AnonMethod-SAC w/o Aux, AnonMethod-SAC, and their ProDVI-enhanced variants. The aggregate curve reports the mean Deep-SAC-normalized score across the 5 Gym tasks. Shaded areas indicate 95\% bootstrap confidence intervals.
}
\label{curves3}
\end{figure*}

This section reports the per-task results and learning curves for the main experiments. Tables~\ref{full_gym_result} and~\ref{full_dmc_result} report the results of AnonMethod, AnonMethod w/o Aux, and their ProDVI-enhanced variants on Gym and DMControl, respectively. Table~\ref{sac_gym} reports the results of AnonMethod-SAC, AnonMethod-SAC w/o Aux, and their ProDVI-enhanced variants on Gym. Tables~\ref{robustness_gym_full} and~\ref{robustness_dmc_full} report the results obtained using two additional independently generated dynamics-prior programs on Gym and DMControl, respectively. Figures~\ref{curves1} and~\ref{curves2} show the learning curves of AnonMethod, AnonMethod w/o Aux, and their ProDVI-enhanced variants on Gym and DMControl, respectively. Figure~\ref{curves3} shows the learning curves of AnonMethod-SAC, AnonMethod-SAC w/o Aux, and their ProDVI-enhanced variants on Gym.

\begin{table*}[t]
    \centering
    \begin{tabular}{lcccc}
        \toprule
        \textbf{Task} & \textbf{\makecell{AnonMethod \\ w/o Aux}} & \textbf{\makecell{AnonMethod \\ w/o Aux + ProDVI}} & \textbf{AnonMethod} & \textbf{\makecell{AnonMethod \\ + ProDVI}} \\
        \midrule
        \multicolumn{5}{l}{\textit{0.25M}} \\
        \midrule     
        Ant-v5 & 3472 [2495, 4519] & 
        5867 [5303, 6376] &
        4621 [3331, 5911] & 
        5662 [4521, 6417] \\
        
        HalfCheetah-v5 & 11799 [11424, 12218] & 
        12412 [12178, 12620] &
        12330 [11811, 12738] & 
        12619 [12395, 12913] \\        

        Hopper-v5 & 1193 [880, 1660] & 
        2372 [1693, 3091] &
        1550 [1060, 2040] & 
        2883 [1574, 3364] \\ 

        Humanoid-v5 & 4578 [3575, 5794] & 
        6178 [5608, 6719] &
        6281 [4952, 7254] & 
        7109 [6556, 7569] \\ 

        Walker2d-v5 & 2942 [2209, 3676] & 
        3018 [2030, 3943] &
        2199 [1590, 3181] & 
        4001 [3256, 4946] \\ 

        \midrule
        \multicolumn{5}{l}{\textit{0.50M}} \\
        \midrule     
        Ant-v5 & 5868 [5290, 6420] & 
        6765 [6032, 7310] &
        5610 [3504, 7142] & 
        7317 [6866, 7700] \\
        
        HalfCheetah-v5 & 13774 [13447, 14213] & 
        14114 [13824, 14426] &
        14387 [14084, 14751] & 
        15874 [14112, 16541] \\        

        Hopper-v5 & 1897 [1379, 2645] & 
        2116 [1598, 2846] &
        1874 [1776, 1990] & 
        2419 [1607, 3231] \\ 

        Humanoid-v5 & 7810 [7307, 8313] & 
        9107 [8909, 9260] &
        9201 [8921, 9455] & 
        9039 [8555, 9484] \\ 

        Walker2d-v5 & 4384 [4024, 4733] & 
        5928 [5316, 6498] &
        4169 [2339, 5458] & 
        5967 [5544, 6389] \\ 

        \midrule
        \multicolumn{5}{l}{\textit{1.00M}} \\
        \midrule     
        Ant-v5 & 6745 [5862, 7453] & 
        7804 [7527, 7978] &
        8006 [7552, 8454] & 
        8369 [7562, 9177] \\
        
        HalfCheetah-v5 & 16597 [16336, 16859] & 
        16587 [16435, 16870] &
        16794 [16622, 17038] & 
        17098 [16806, 17391] \\        

        Hopper-v5 & 1701 [1582, 1861] & 
        2431 [1589, 2933] &
        2477 [2207, 2883] & 
        2758 [2103, 3414] \\ 

        Humanoid-v5 & 9259 [8853, 9665] & 
        9986 [9585, 10369] &
        10022 [9919, 10141] & 
        10118 [9574, 10410] \\ 

        Walker2d-v5 & 4653 [3318, 5747] & 
        6101 [4948, 7058] &
        6048 [5274, 6724] & 
        6457 [6096, 6830] \\ 
        \bottomrule
    \end{tabular}
    \caption{
    Per-task results on the Gym benchmark for AnonMethod w/o Aux, AnonMethod, and their ProDVI-enhanced variants. We report raw episode returns at 0.25M, 0.50M, and 1.00M environment steps.
    % Results are averaged over 5 seeds. 
    The {{[bracketed values]}} represent a 95\% bootstrap confidence interval.
    }
    \label{full_gym_result}
\end{table*}

\begin{table*}[t]
    \centering
    \begin{tabular}{lcccc}
        \toprule
        \textbf{Task} & \textbf{\makecell{AnonMethod \\ w/o Aux}} & \textbf{\makecell{AnonMethod \\ w/o Aux + ProDVI}} & \textbf{AnonMethod} & \textbf{\makecell{AnonMethod \\ + ProDVI}} \\
        \midrule
        \multicolumn{5}{l}{\textit{0.25M}} \\
        \midrule     
        dog-run & 207 [166, 259] & 
        222 [193, 262] &
        170 [161, 181] & 
        162 [154, 170] \\
        
        dog-stand & 885 [861, 905] & 
        869 [818, 909] &
        862 [842, 882] & 
        938 [919, 953] \\        

        dog-trot & 254 [210, 297] & 
        230 [178, 279] &
        250 [225, 276] & 
        224 [198, 251] \\ 

        dog-walk & 354 [318, 396] & 
        483 [411, 563] &
        344 [255, 679] & 
        546 [451, 900] \\ 

        humanoid-run & 78 [37, 109] & 
        96 [48, 128] &
        113 [109, 117] & 
        123 [119, 128] \\ 

        humanoid-stand & 372 [323, 421] & 
        536 [390, 686] &
        556 [463, 691] & 
        549 [516, 574] \\ 

        humanoid-walk & 361 [305, 425] & 
        338 [162, 453] &
        459 [435, 483] & 
        491 [465, 524] \\ 

        \midrule
        \multicolumn{5}{l}{\textit{0.50M}} \\
        \midrule     
        dog-run & 414 [346, 481] & 
        426 [361, 491] &
        296 [273, 316] & 
        336 [294, 396] \\
        
        dog-stand & 936 [903, 962] & 
        956 [937, 971] &
        933 [924, 943] & 
        964 [960, 972] \\        

        dog-trot & 582 [479, 695] & 
        589 [429, 748] &
        738 [605, 818] & 
        706 [597, 787] \\ 

        dog-walk & 813 [755, 854] & 
        865 [847, 886] &
        803 [710, 866] & 
        886 [866, 900] \\ 

        humanoid-run & 147 [140, 156] & 
        154 [146, 162] &
        172 [161, 189] & 
        170 [162, 181] \\ 

        humanoid-stand & 728 [632, 824] & 
        782 [656, 881] &
        836 [735, 897] & 
        856 [815, 898] \\ 

        humanoid-walk & 504 [481, 531] & 
        565 [495, 696] &
        615 [598, 630] & 
        601 [575, 623] \\ 

        \bottomrule
    \end{tabular}
    \caption{
    Per-task DMControl results for AnonMethod w/o Aux, AnonMethod, and their ProDVI-enhanced variants. We report raw episode returns on the dog and humanoid tasks at 0.25M and 0.50M environment steps.
    % Results are averaged over 5 seeds. 
    The {{[bracketed values]}} represent a 95\% bootstrap confidence interval.
    }
    \label{full_dmc_result}
\end{table*}

\begin{table*}[t]
    \centering
    \begin{tabular}{lcccc}
        \toprule
        \textbf{Task} & \textbf{\makecell{AnonMethod-SAC \\ w/o Aux}} & \textbf{\makecell{AnonMethod-SAC \\ w/o Aux + ProDVI}} & \textbf{\makecell{AnonMethod-SAC}} & \textbf{\makecell{AnonMethod-SAC \\ + ProDVI}}  \\
        \midrule
        \multicolumn{5}{l}{\textit{0.25M}} \\
        \midrule     
        Ant-v5 & 2375 [936, 3838] & 
        5069 [4132, 5927] &
        4855 [4451, 5281] & 
        5106 [3783, 6429] \\
        
        HalfCheetah-v5 & 10813 [10110, 11282] & 
        10751 [9858, 11634] &
        10613 [9756, 11532] & 
        12118 [11790, 12460] \\        

        Hopper-v5 & 1973 [1235, 2734] & 
        2253 [1533, 2977] &
        1952 [1466, 2641] & 
        2326 [1606, 3046] \\ 

        Humanoid-v5 & 1624 [786, 2462] & 
        1858 [1212, 2641] &
        2435 [1373, 3765] & 
        6769 [4342, 8400] \\ 

        Walker2d-v5 & 2190 [786, 3618] & 
        2515 [1954, 3141] &
        1550 [428, 2673] & 
        3783 [2017, 5089] \\ 

        \midrule
        \multicolumn{5}{l}{\textit{0.50M}} \\
        \midrule     
        Ant-v5 & 5394 [4011, 6421] & 
        6551 [5830, 7040] &
        6689 [5770, 7258] & 
        6775 [6129, 7363] \\
        
        HalfCheetah-v5 & 12739 [12142, 13256] & 
        13292 [12603, 14099] &
        12901 [12296, 13396] & 
        14248 [13266, 15231] \\        

        Hopper-v5 & 2953 [1970, 3745] & 
        2658 [1937, 3313] &
        2115 [1567, 2889] & 
        2520 [1864, 3327] \\ 

        Humanoid-v5 & 6071 [3794, 7940] & 
        5605 [2460, 8750] &
        5705 [3355, 7373] & 
        7991 [6225, 9758] \\ 

        Walker2d-v5 & 3282 [2378, 4179] & 
        4172 [2995, 5436] &
        3102 [2454, 3698] & 
        4684 [3534, 5834] \\ 

        \midrule
        \multicolumn{5}{l}{\textit{1.00M}} \\
        \midrule     
        Ant-v5 & 5733 [3992, 7425] & 
        7620 [7396, 7828] &
        7544 [6854, 7965] & 
        7441 [6144, 8234] \\
        
        HalfCheetah-v5 & 15861 [15580, 16143] & 
        15895 [15270, 16407] &
        15759 [15356, 16133] & 
        16512 [16305, 16707] \\        

        Hopper-v5 & 2509 [2064, 2990] & 
        2332 [1950, 2771] &
        2165 [1727, 2621] & 
        2639 [1979, 3358] \\ 

        Humanoid-v5 & 7364 [4798, 8968] & 
        6707 [3833, 8964] &
        8120 [7248, 9140] & 
        8043 [6539, 9464] \\ 

        Walker2d-v5 & 4062 [2669, 5482] & 
        5445 [4783, 6108] &
        6047 [5355, 6604] & 
        6592 [5212, 7972] \\ 
        \bottomrule
    \end{tabular}
    \caption{
    Per-task Gym results for the SAC-based variants, including AnonMethod-SAC w/o Aux, AnonMethod-SAC, and their ProDVI-enhanced counterparts. We report raw episode returns at 0.25M, 0.50M, and 1.00M environment steps. Results are averaged over 5 seeds. The {{[bracketed values]}} represent a 95\% bootstrap confidence interval.
    }
    \label{sac_gym}
\end{table*}

\begin{table*}[t]
    \centering
    \begin{tabular}{lcc}
        \toprule
        \textbf{Task} & \textbf{{Dynamics Prior \# 2}} & \textbf{Dynamics Prior \# 3} \\
        \midrule
        \multicolumn{3}{l}{\textit{0.25M}} \\
        \midrule     
        Ant-v5 & 5129 [4258, 5920] & 
        4674 [3442, 5771] \\
        
        HalfCheetah-v5 & 12466 [12129, 12803] & 
        12525 [12257, 12957] \\        

        Hopper-v5 & 2239 [1435, 3042] & 
        2324 [1594, 3053] \\ 

        Humanoid-v5 & 6684 [5458, 7799] & 
        7056 [6428, 7683] \\ 

        Walker2d-v5 & 3034 [2055, 4013] & 
        4027 [2771, 5088] \\ 

        \midrule
        \multicolumn{3}{l}{\textit{0.50M}} \\
        \midrule     
        Ant-v5 & 7103 [6211, 7601] & 
        6403 [5423, 7398] \\
        
        HalfCheetah-v5 & 14244 [13691, 14797] & 
        13569 [12475, 14464] \\        

        Hopper-v5 & 2781 [2216, 3344] & 
        2500 [1711, 3174] \\ 

        Humanoid-v5 & 9091 [8846, 9343] & 
        9418 [9075, 9695] \\ 

        Walker2d-v5 & 4388 [2969, 5816] & 
        5390 [4203, 6422] \\ 

        \midrule
        \multicolumn{3}{l}{\textit{1.00M}} \\
        \midrule     
        Ant-v5 & 8493 [8002, 8982] & 
        7801 [7340, 8146] \\
        
        HalfCheetah-v5 & 16836 [16512, 17160] & 
        17057 [16930, 17133] \\        

        Hopper-v5 & 2710 [2068, 3351] & 
        2385 [1931, 2843] \\ 

        Humanoid-v5 & 10194 [10032, 10347] & 
        10362 [10266, 10459] \\ 

        Walker2d-v5 & 5738 [4857, 6294] & 
        6558 [5994, 7207] \\ 
        
        \bottomrule
    \end{tabular}
    \caption{ Per-task evaluation returns of AnonMethod with ProDVI on Gym using two independently generated dynamics-prior programs, Dynamics Prior \#2 and Dynamics Prior \#3. Results are reported at 0.25M, 0.50M, and 1.00M environment steps. Each entry shows the mean return over 5 seeds, with brackets denoting 95\% bootstrap confidence intervals. }
    \label{robustness_gym_full}
\end{table*}

\begin{table*}[t]
    \centering
    \begin{tabular}{lcc}
        \toprule
        \textbf{Task} & \textbf{{Dynamics Prior \# 2}} & \textbf{Dynamics Prior \# 3} \\
        \midrule
        \multicolumn{3}{l}{\textit{0.25M}} \\
        \midrule     
        dog-run & 182 [163, 201] & 
        161 [152, 170] \\
        
        dog-stand & 852 [804, 893] & 
        871 [835, 902] \\        

        dog-trot & 264 [233, 295] & 
        302 [245, 360] \\ 

        dog-walk & 517 [456, 587] & 
        386 [329, 443] \\ 

        humanoid-run & 137 [130, 142] & 
        130 [124, 136] \\ 

        humanoid-stand & 617 [531, 717] & 
        546 [455, 614] \\ 

        humanoid-walk & 495 [467, 523] & 
        455 [419, 493] \\ 

        \midrule
        \multicolumn{3}{l}{\textit{0.50M}} \\
        \midrule     
        dog-run & 392 [353, 430] & 
        314 [257, 367] \\
        
        dog-stand & 952 [938, 965] & 
        953 [948, 959] \\        

        dog-trot & 688 [578, 785] & 
        763 [675, 850] \\ 

        dog-walk & 713 [366, 888] & 
        847 [817, 874] \\ 

        humanoid-run & 174 [166, 181] & 
        198 [167, 232] \\ 

        humanoid-stand & 898 [874, 913] & 
        892 [877, 907] \\ 

        humanoid-walk & 744 [682, 812] & 
        598 [553, 656] \\ 

        \bottomrule
    \end{tabular}
    \caption{ Per-task aggregated returns of AnonMethod with ProDVI on DMControl using two independently generated dynamics-prior programs, Dynamics Prior \#2 and Dynamics Prior \#3. Results are reported at 0.25M and 0.50M environment steps. Each entry shows the mean return over 5 seeds, with brackets denoting 95\% bootstrap confidence intervals. }
    \label{robustness_dmc_full}
\end{table*}

{
\small
\bibliography{aaai2027}}